%% file: main.tex
\PassOptionsToPackage{dvipsnames}{xcolor}
\documentclass[multiauthors,onecolumn,cfinal]{LaTeXclass/cohere}

\usepackage{lipsum}
\usepackage{pdfpages}
\usepackage{colortbl}
\usepackage{tabulary}
\usepackage{longtable}
\usepackage{array}
\usepackage{placeins}
\usepackage{tablefootnote}
\usepackage{tcolorbox}
\usepackage{wrapfig}
\usepackage{breqn} 

\usepackage{pifont}
\definecolor{deeppurple}{HTML}{9e02f7}
\definecolor{forestgreen}{HTML}{2e7d43}
\usepackage{multirow}
\usepackage{tabularx}

\usepackage[utf8]{inputenc}
\usepackage{listings}

\lstdefinestyle{custom}{
    basicstyle=\ttfamily,
    breaklines=true,
    postbreak=\mbox{\textcolor{red}{$\hookrightarrow$}\space},
    showstringspaces=false,
}
\usepackage{arydshln}
\usepackage{booktabs}
\usepackage{multirow}

\usepackage{xspace} 
\usepackage{makecell}
\usepackage{multirow}

\usepackage[framemethod=TikZ]{mdframed}
\usepackage{multicol}
\usepackage{ragged2e}
\usepackage{nicematrix}
\usepackage{comment}
\usepackage{amssymb}
\usepackage{xcolor}

\usepackage{flafter}
\setcitestyle{number,square}

\title{The Art of Asking:}
\date{October 22, 2025}
\subtitle{Multilingual Prompt Optimization for Synthetic Data}

\author{name={David Mora\fa},affiliation={1}}
\author{name={Viraat Aryabumi},affiliation={2}}
\author{name={Wei-Yin Ko},affiliation={2}}
\author{name={Sara Hooker},affiliation={1}}
\author{name={Julia Kreutzer},affiliation={1}}
\author{name={Marzieh Fadaee},affiliation={1}}
\affiliations{
    \item[1] Cohere Labs
    \item[2] Cohere
}

\corresponding[*]{\{\url{david.mora}, \url{juliakreutzer}, \url{marzieh}\}\url{{@cohere.com}}}

\abstract{
\justifying
Synthetic data has become a cornerstone for scaling LLMs, yet its benefits for multilingual modeling remain bottlenecked by translation.
Translation for data augmentation inherits English-centric framing and style and neglects cultural dimensions, ultimately constraining model generalization.
We argue that the overlooked prompt space offers a more powerful lever for improving multilingual performance.
We introduce a lightweight framework for prompt-space optimization, where translated prompts are systematically transformed for \textit{Naturalness}, \textit{Cultural Adaptation,} and \textit{Difficulty Enhancement}. 
Using an off-the-shelf multilingual LLM, we apply these transformations to prompts from 12 languages spanning 7 families.
Under identical data conditions, our approach achieves substantial and consistent downstream improvements over the translation-only baseline: +4.7\% accuracy on Global-MMLU, +2.4\% XCometXL on Flores  and +35.3\% wins in preferences on mArenaHard. These gains are further corroborated by human preferences on open-ended creative writing.
We establish prompt-space optimization as a simple yet powerful paradigm for building multilingual LLMs that are more robust, culturally grounded, and globally capable.
}

\renewcommand{\FONTauthors}{\bfseries\large}

\begin{document}

\section{Introduction}

The field of synthetic data generation has largely operated under a \textbf{generation-focused paradigm}: given existing prompts, optimize the quality of the generated completions~\citep{long-etal-2024-llms,liu2024best}, via e.g. targeted filtering~\citep{grattafiori2024llama3herdmodels,shimabucoro-etal-2024-llm} or test-time scaling~\citep{muennighoff2025s1simpletesttimescaling}.
However, completions are only as diverse and representative as the inputs they are conditioned on, and numerous studies show that prompts themselves can often be noisy or low quality leading synthetic data to reinforce these deficiencies rather than systematically broadening the training distribution \citep{schreiter2025promptengineeringpromptvocabulary,he2024doespromptformattingimpact}.

This challenge is especially acute in the multilingual setting, where translation-based prompt expansion dominates instruction tuning~\citep{ustun-etal-2024-aya,dang2024ayaexpansecombiningresearch,chen-etal-2024-good-data, martins2025eurollm9btechnicalreport}.
While effective for scaling coverage, translations introduce artifacts such as unnatural phrasing (\textit{translationese})~\citep{10.1162/COLI_a_00111,eetemadi-toutanova-2014-asymmetric}, lexical errors, or shifts in toxicity~\citep{ermis-etal-2024-one}.
Even high-quality translations project the semantics of the original English prompts, but rarely adapt content for cultural relevance~\citep{enomoto-etal-2025-fair}. This perpetuates an English-centric perspective: models are optimized for many target languages, but still trained on prompts that reflect the needs, assumptions, and discourse patterns of English speakers.
Prior work shows that this mismatch has measurable downstream effects on both generation quality and fairness~\citep{li-etal-2025-lost}.

\begin{figure*}[t]
    \centering
    \includegraphics[width=\textwidth,trim=0 44 0 62,clip]{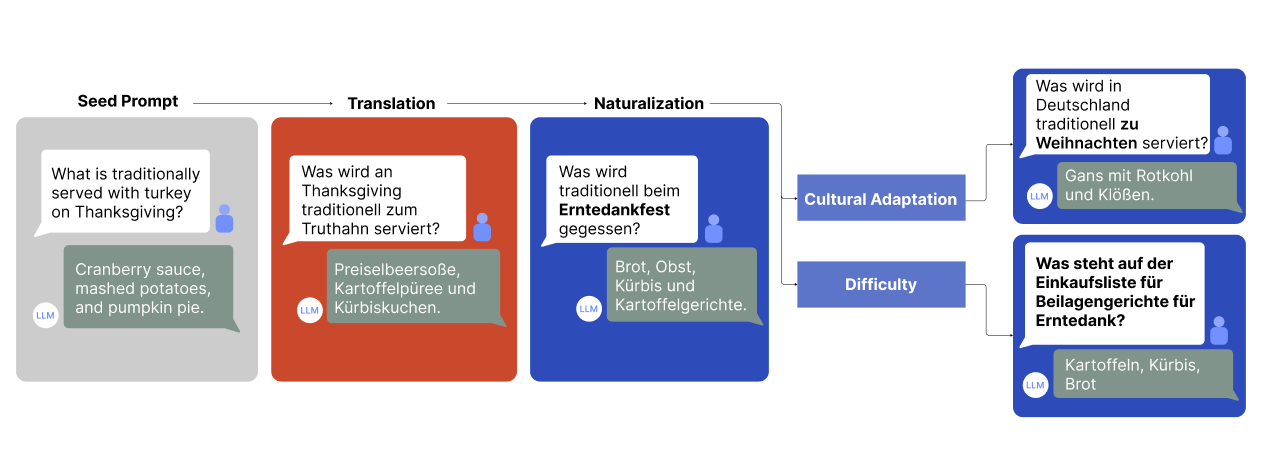}
    \caption{\textbf{Illustration of our prompt transformations on a representative toy example that gets adapted for German:} Each transformation modifies the original English prompt, with major modifications highlighted in bold. Modifications to the prompt cause changes in the generation as well, so by making the prompt more natural by using the German term ``Erntedankfest'' rather than the English ``Thanksgiving'', the completion now lists typical German rather than American Thanksgiving dishes (``bread, fruit, pumpkin, potato dishes''). The \textit{Cultural Adaptation}further localizes the prompt (``in Germany'') and replaces the event of Thanksgiving with the event of Christmas, which has larger significance in German culture. The \textit{Difficulty} transformation yields a prompt that requests a shopping list for side dishes of Thanksgiving, making it more specific but also more complex. Full examples of prompt transformations and their corresponding completions that were used for our experiments are in Table \ref{app-tab:examples-table}.} 
        \label{fig:transformation}
\end{figure*}

In this paper, we introduce a \textbf{prompt-focused paradigm} for synthetic data generation, where translated prompts are systematically transformed along three critical dimensions: \textit{Naturalness}, \textit{Cultural Adaptation}, and \textit{Difficulty Enhancement}.
By treating prompts as dynamic components rather than fixed scaffolds, we directly reshape the input distribution, reducing translation artifacts and embedding inductive biases that are better aligned with real user data (\Cref{fig:transformation}).

\begin{wrapfigure}{r}{0.48\textwidth}
\centering
\vspace{-0.8\baselineskip}
\includegraphics[width=\linewidth]{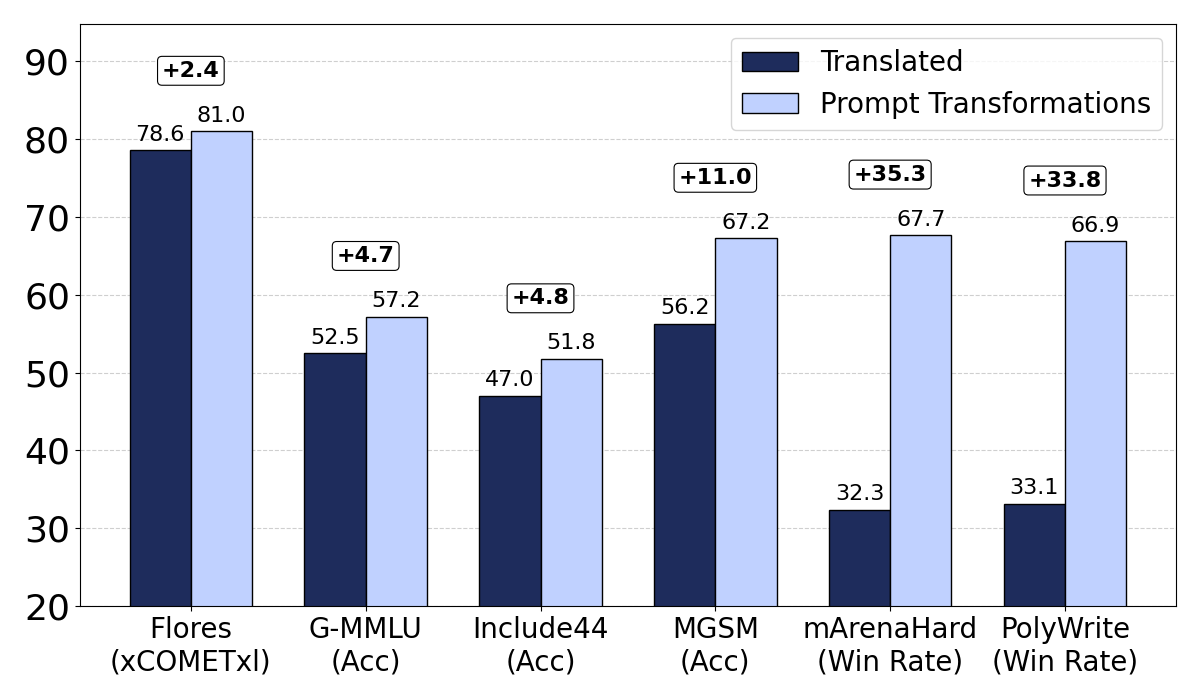}
\caption{\textbf{Prompt transformations outperform translation-only prompts.} We compare the translated baseline with our \textit{Cultural+Difficulty Mix} across multilingual benchmarks. mArenaHard and PolyWrite are direct pairwise wins.}
\label{fig:main_results}
\end{wrapfigure}
Starting from translated English prompts, we apply targeted prompt-space transformations using a strong teacher LLM, and measure their impact both on the data itself and on downstream performance. We evaluate this approach across 12 languages spanning diverse families and find that our prompt transformations \textbf{successfully improve quality along the targeted dimensions}: \textit{Naturalness} increases lexical diversity, \textit{Cultural Adaptation} enhances fluency, and the \textit{Difficulty Enhancement} transformation raises both difficulty and overall quality (at the cost of diversity) when compared to translated prompts. These prompt-side improvements  carry over to completions where \textbf{even small interventions in the prompts lead to substantial changes in completions} (\Cref{tab:metrics_prompt_completion}). Finally, when used for fine-tuning a 7B base model, these effects yield strong and \textbf{consistent improvements across all languages and a diverse set of benchmarks} (mathematical reasoning, translation, language and culture understanding, open-ended generation) with particularly pronounced gains on open-ended tasks (\Cref{fig:main_results}), also confirmed with human evaluation on a subset of languages.

As our experiments show, bootstrapping fine-tuning data from translations via targeted transformations has a tremendous impact on the state of language modeling especially for languages that are typically overlooked in LLM development.

\section{Method}\label{sec:method}

\enlargethispage{2\baselineskip}
Existing synthetic data pipelines primarily expand $P(y \mid x)$, the conditional mapping from prompts to completions, while implicitly assuming that the input prompt distribution $P(x)$ is fixed. 
This \textit{generation-focused} view limits diversity and cultural grounding: completions remain tied to the artifacts, biases, and topical scope of the original prompts, especially when these are machine-translated from English. 
We instead intervene directly on the input distribution $P(x)$, introducing an inductive bias toward more natural, contextually grounded, and linguistically rich prompts. 
This \textit{prompt-focused} perspective reframes synthetic data generation as optimization in the \textbf{prompt space}, not just in the \textbf{generation space}.

\subsection{Problem Setup}
Let $P_{\text{src}}(x)$ denote the distribution of prompts in a high-resource source language (e.g., English). 
We yield a corresponding target-language distribution $P_{\text{trg},\ell}(x)$ for each language $\ell$ through translation:
\[
x^{\text{trg}} \sim P_{\text{trg},\ell} = \text{translate}(P_{\text{src}}).
\]
While this step expands coverage, it does not adapt content to the linguistic or cultural norms of the target language.
We therefore introduce a transformation operator $\mathcal{T}$ that refines translated prompts:
\[
x^{\text{opt}} = \mathcal{T}(x^{\text{trg}}), \quad x^{\text{opt}} \sim P_{\text{opt},\ell}.
\]
The resulting optimized distribution $P_{\text{opt},\ell}$ replaces $P_{\text{trg},\ell}$ as the input space for training, giving rise to
\[
P_{\text{train},\ell}(x, y) = P_{\text{opt},\ell}(x) \, P_{\text{teacher}}(y \mid x).
\]
In this setup, any shift in $P_{\text{opt},\ell}$ directly influences the inductive bias of the fine-tuned model, altering not only what it learns to \textit{say} ($P(y \mid x)$), but also what it learns to \textit{understand}.

\subsection{Transformation Operators}
We instantiate $\mathcal{T}$ as a family of modular operators 
$\mathcal{T} = \{\mathcal{T}_{\text{nat}}, \mathcal{T}_{\text{cult}}, \mathcal{T}_{\text{diff}}\}$,
each targeting a distinct dimension of prompt quality:

\begin{itemize}[leftmargin=*]
    \item \textbf{Naturalness} ($\mathcal{T}_{\text{nat}}$): Removes translation artifacts and restores idiomatic phrasing to better reflect authentic language use.
    \item \textbf{Cultural adaptation} ($\mathcal{T}_{\text{cult}}$): Recontextualizes prompts to locally relevant examples, values, and references, aligning them with cultural norms.
    \item \textbf{Difficulty enhancement} ($\mathcal{T}_{\text{diff}}$): Increases task complexity by expanding or reformulating prompts into more challenging instructions.
\end{itemize}

Each transformation produces a valid optimized prompt distribution $P_{\text{opt},\ell}$; in practice, these operators can be applied individually or in sequence (e.g., \textit{Naturalness} followed by \textit{Cultural Adaptation}).
Each operator shifts $P_{\text{opt},\ell}$ closer to the true user distribution $P^{*}_{\ell}$, improving both data quality and downstream generalization.

\subsection{Prompt Transformations via LLMs}
Each transformation $\mathcal{T}$ is executed with an LLM. At the core of the transformation is a prompt that specifies which context and input (e.g. user prompt, original English prompt, target language) is included in the transformation, its description and some additional guidelines (the exact prompt templates are given in \cref{tab:prompts}). 
These were improved over a few iterations via manual data inspection, but they can be further customized for other domains.  We kept the prompts simple as overly rigid guidelines risk reducing diversity, making outputs feel templated, limiting generalization and correctness, especially in underrepresented languages where the teacher model may already struggle with instruction following and hallucinate more easily.

\section{Experiments}
We set up a multilingual fine-tuning pipeline where the primary goal is to improve quality and performance in various tasks, with special focus on naturalness and fluency of open-ended generations, cultural adequacy and accuracy in challenging domains that typically show strong language disparities. 
The setup aims to make the impact of each of our transformations measurable, first in the resulting \textit{data}, and then further in \textit{downstream performance} of the model. 

\subsection{Data Processing Pipeline}

\subsubsection{English Seed Prompts} We collect real prompts from users around the world (with consent and not including PII), similar to ShareLM~\citep{don-yehiya-etal-2025-sharelm}. Because the prompts are noisy, we apply content filtering and language identification filtering with FastText \citep{joulin2016fasttext,joulin2016bag} to extract 187k English prompts. This pool of prompts is attractive for modeling because these are unseen samples of real-life use with state-of-the-art models, and thereby provide an excellent learning opportunity.
\subsubsection{Prompt Translation into Target Languages} We take distinct 10k sub-samples from the English pool of prompts and automatically translate them into 12 target languages (listed in \cref{tab:languages}), using an in-house state-of-the-art translation model.

While geographically close (all spoken in Europe), these languages cover seven language families (including one isolate, Basque) and three scripts. They are standardized and have mid to high institutional support~\citep{bird-2022-local}, but vary in terms of their availability of accessible, high-quality data, representation on the web and in NLP research, and support in open LLMs~\citep{ranathunga-de-silva-2022-languages}. 
As a result (\cref{tab:languages}), the translation capabilities of our expert translation model varies, yielding top quality e.g. for German, Spanish and Czech, but much poorer quality e.g. for Latvian, Basque and Welsh. The translation quality on our domain of user-submitted prompts is overall slightly lower (but also harder to estimate), due to challenging inputs like code or non-standard language.
Nevertheless, we assume that for this selection of languages, bootstrapping with translation and transformation is feasible, and it lets us study our proposed methods on a diverse spectrum.

\subsubsection{Prompt Optimization} We choose \textsc{Gemma3-27B-it}\footnote{\url{https://huggingface.co/google/gemma-3-27b-it}} as our transformation model for its broad language support and strong multilingual performance~\citep{gemmateam2025gemma3technicalreport} (see translation evaluation in \cref{tab:languages}).
For each transformation described in \cref{sec:method}, we prompt it with the respective custom instruction and sample a single generation with a temperature of 0.3. 
Importantly, we apply the \textit{Naturalness} transformation directly to the translated prompts, but for the \textit{Cultural Adaptation} and \textit{Difficulty Enhancement} transformations, we apply them on top of the \textit{Naturalness}-transformed prompts. This decision is based on our initial experiments, which showed that the \textit{Naturalness} transformation provides a mild, generally beneficial adjustment that does not interfere with the other two.
After transforming the prompts, we filter the data with FastText's language identification to prevent language confusion~\citep{marchisio-etal-2024-understanding}.

\subsubsection{Prompt Completions} To generate completions, we rely on a teacher model without any additional instructions. For this purpose, we use the same model as our transformation model, \texttt{Gemma3-27B-IT}.\footnote{In principle both models do not need to be identical, it is a choice of convenience.}
For each prompt, we sample a single generation with a temperature of 0.3. To ensure that outputs are produced in the intended language, we once again run language identification and discard mismatches (language counts can be found in \cref{tab:language_counts}). We adopt this simple completion generation setup in order to cleanly isolate the effect of our prompt interventions.

\subsection{Fine-Tuning}

\subsubsection{Base Model}\label{model} We use 
the base version of CommandR7B,\footnote{\url{https://docs.cohere.com/docs/command-r7b}} an open-weights 7B multilingual model 
pre-trained on the following 23 languages: English, French, Spanish, Italian, German, Portuguese, Japanese, Korean, Arabic, Chinese, Russian, Polish, Turkish, Vietnamese, Dutch, Czech, Indonesian, Ukrainian, Romanian, Greek, Hindi, Hebrew, and Persian. 
Only five of these languages overlap with our target languages (see \cref{tab:languages}), which enables us to study the effectiveness of our transformation techniques in expanding the language coverage of LLMs during post-training (\cref{sec:analysis}). 
Supervised fine-tuning (SFT) follows a standard procedure, details described in \cref{app:training}.

\subsubsection{Data Mixture} 
We consider four main datasets, one for each of the transformations described in \cref{sec:method} and an additional one where we mix 50\% of \textit{Culturally Adapted} data and 50\% of the \textit{Difficulty Enhanced} data. 
 We complement our datasets with a portion of other standard instruction tuning datasets (mostly English) in order to reduce overfitting, these include domains like math, code, reasoning but also multilingual datasets (for the 23 languages supported by the base model).
In total, each of our four data mixtures contains roughly 590k examples, around 48\% of which are contributed by our prompt transformed datasets (counts in \cref{tab:language_counts}).

\subsection{Evaluation}
Our primary question throughout will be how our transformations compare against only prompt translation. We compare this in two stages: at the data level, and in downstream evaluations.

\subsubsection{Data Evaluations} \label{data_evals} We evaluate the textual characteristics of both prompts and completions using a combination of standard metrics and LLM-based scores.
First, we measure how much the prompts and generations have changed in comparison to their translated counterparts at the surface level, using relative edit distance (Levenshtein distance normalized by the maximum length) and length in characters.
To assess diversity, we compute corpus n-gram diversity at the language level by tokenizing the texts using spaCy\footnote{\url{https://spacy.io/}} and then computing the ratio of unique n-grams to total n-grams~\citep{padmakumar2024doeswritinglanguagemodels,shaib2025standardizingmeasurementtextdiversity}. To assess naturalness, we use \textsc{Gemma3-27b-pt} to compute the perplexity of each text. Previous works have used target language model perplexity as a metric for \textit{translationese}~\citep{bizzoni-lapshinova-koltunski-2021-measuring, li-etal-2025-lost}. To assess quality and difficulty, we rely on automatic scoring by prompting an LLM (\textsc{Gemma3-27b-it})
to score the texts on a discrete scale (prompts included in \cref{app:eval-prompts}). These measures indicate whether our transformations succeed in eliciting more desirable textual features which are key to steering downstream performance~\citep{shimabucoro-etal-2024-llm}.

\subsubsection{Downstream evaluations}

\textbf{Discriminative benchmarks.} Our suite covers two discriminative multi-choice tasks: Include44~\citep{romanou2024includeevaluatingmultilinguallanguage}, with questions from local academic and professional exams in target languages, and Global-MMLU (G-MMLU)~\citep{singh-etal-2025-global} with translated QA tasks from English. We expect these tasks can help us measure language disparities in knowledge access, especially for questions that are culture-specific. Implementation details are in \cref{app:evals}.

\textbf{Close-ended generative benchmarks.} For these benchmarks, there exist gold standard outputs which quality can be measured against. This is interesting because it allows us to precisely track quality improvements (as in discriminative benchmarks), but also captures the quality of more than one output tokens (as opposed to discriminative benchmarks).  
We choose the Flores translation task~\citep{nllb2022} for its wide language coverage, and MGSM~\citep{shi2023language} as a challenging multilingual math task (especially for the lower-resourced languages). 
For MGSM, we extend the original language coverage by adding translated versions, which we refer to as MGSM++ (the sources of these translations are listed in \cref{app:evals}).

\textbf{Open-ended generative benchmarks.} Our primary target are the following two benchmarks that capture open-ended generation quality:\footnote{Prior work found discriminative benchmarks not indicative enough for generative performance~\citep{ustun-etal-2024-aya}.} 
 m-ArenaHard v2.0~\citep{khairi2025lifegivessamplesbenefits} is a collection of challenging LMArena prompts~\citep{zheng2024lmsyschatm} that was translated into 23 languages. It contains prompts from a wide range of domains, but especially code and math. 
 We extend the set of support languages to include our missing ones, by translating the prompts from English (and apply language filtering to the prompts), forming mArenaHard++ v2.0 (the same procedure as for the original mArenaHard-v2.0).
 Performance is measured with win rates in pairwise comparisons against a competitor model as judged by GPT-4.1 (\textsc{gpt-4.1-2025-04-14}).\footnote{\url{https://platform.openai.com/docs/models/gpt-4.1}}
 To capture language naturalness better, we compare our models on creative writing prompts from the PolyWrite benchmark~\citep{ji2024emma500enhancingmassivelymultilingual}, where we additionally compute win-rates with a judge prompt that evaluates the naturalness of completions, and evaluate the diversity of the generations with self-BLEU~\citep{10.1145/3209978.3210080,ji2024emma500enhancingmassivelymultilingual}.

\textbf{Language coverage.} Although not all target languages are covered in GlobalMMLU, Include44 and MGSM++, each language is represented in at least one of them, while all being included in the remaining, see \cref{tab:eval_suite}.
We only evaluate the models for our focus languages (plus English, where available), and report averages (plus breakdowns in \cref{app:breakdown}).

\section{Results}

\subsection{Data Quality}

\begin{table*}[h!]
\centering
\setlength{\tabcolsep}{6pt}
\resizebox{0.9\textwidth}{!}{
\begin{tabular}{%
  l
  cc  
  cc  
  cc  
  cc  
  cc  
  cc  
}
\toprule
 & \multicolumn{2}{c}{\textbf{Length}} & \multicolumn{2}{c}{\textbf{Rel.\ Dist.$\uparrow$}} & \multicolumn{2}{c}{\textbf{Perplexity}$\downarrow$}
 & \multicolumn{2}{c}{\textbf{Diversity$\uparrow$}} & \multicolumn{2}{c}{\textbf{Difficulty$\uparrow$}} & \multicolumn{2}{c}{\textbf{Quality$\uparrow$}} \\
\cmidrule(lr){2-3}\cmidrule(lr){4-5}\cmidrule(lr){6-7}\cmidrule(lr){8-9}\cmidrule(lr){10-11}\cmidrule(lr){12-13}
\textbf{Transformation}
  & {P} & {C}
  & {P} & {C}
  & {P} & {C}
  & {P} & {C}
  & {P} & {C}
  & {P} & {C} \\
\midrule
Translated
  & 406 & 2451
  & \multicolumn{1}{c}{--} & \multicolumn{1}{c}{--}
  & 15.34 & 2.46
  & 0.88 & 0.77
  & 1.78 & 1.77
  & 3.21 & 4.78 \\

\midrule
Naturalized
  & 397 & 2490
  & 0.24 & 0.64
  & 14.06 & 2.48
  & \textbf{0.90} & 0.77
  & 1.76 & 1.75
  & 3.26 & 4.81 \\

Cultural
  & 470 & 2352
  & 0.30 & 0.67
  & 12.11 & 2.51
  & 0.89 & \textbf{0.79}
  & 1.76 & 1.76
  & 3.28 & 4.82 \\

Difficulty
  & 1936 & 5322
  & \textbf{0.86} & \textbf{0.81}
  & \textbf{3.13} & \textbf{2.15}
  & 0.77 & 0.76
  & \textbf{2.44} & \textbf{2.45}
  & \textbf{4.50} & \textbf{4.83} \\

\hdashline
Cultural + Diff.
  & 1205 & 3873
  & 0.58 & 0.74
  & 4.52 & 2.27
  & 0.82 & 0.77
  & 1.97 & 2.10
  & 3.75 & 4.76 \\
\bottomrule
\end{tabular}%
}
\caption{\textbf{Text metrics for prompts (P) and completions (C).} Lower perplexity and higher diversity, difficulty, and quality are better (1000 prompts per language).}
\label{tab:metrics_prompt_completion}
\end{table*}

\subsubsection{Prompt Quality} \Cref{tab:metrics_prompt_completion} confirms that our transformations advance the quality of the prompts (``P'' columns) over the original translated prompts on all dimensions, in terms of diversity, fluency, and also general quality and difficulty.
The \textit{Naturalness} transformation achieves the greatest n-gram diversity, which confirms that it re-introduces linguistic richness that might have gotten lost in translation. 
The \textit{Cultural Adaptation} transformation significantly lowers perplexity, showing that it is closely aligned to the target-language content that the base model has seen during pretraining.
The \textit{Difficulty} transformation is the most aggressive transformation, as its edit distance from the translated prompts is more than 3$\times$ higher than the other transformations. It also increases the prompt length by an average factor of 4.8$\times$. 
We manually inspect a subset of these prompts and find that the \textit{Difficulty} transformation typically introduces additional constraints, 
which are similar in template across data points, consequently lowering the diversity. 
Our LLM judge considers these prompts as of substantially higher quality than the naturalized or cultural ones. We thus expect the largest impact on generations and downstream from this transformation.
When mixing difficulty and cultural data, we obtain scores in between both individual transformations, which, compared to difficulty alone, raises n-gram diversity, but lowers the other metrics.

\begin{table*}[t]
\centering
\resizebox{\columnwidth}{!}{%
\begin{tabular}{lcccccc}
\toprule
\multirow{2}{*}{\textbf{Prompts in FT}} & \textbf{Flores} & \textbf{G-MMLU} & \textbf{Include44} & \textbf{MGSM} & 
\textbf{mArenaHard} & \textbf{PolyWrite} \\
\cmidrule(lr){2-2} \cmidrule(lr){3-5} \cmidrule(lr){6-7}
 & xCometXL $\uparrow$ & \multicolumn{3}{c}{Accuracy $\uparrow$} & 
\multicolumn{2}{c}{Win-rate \% $\uparrow$} \\
\midrule
Translated & 0.786 & 52.5 & 47.0 & 56.2 & -- & -- \\
\hline
Naturalized & 0.791 & 53.1 & 46.9 & 56.5 & 57.7 & 63.8 \\
Cultural & 0.805 & \textbf{57.9} & 50.8 & 66.0 & 65.7 & 66.1\\
Difficulty & \textbf{0.816} & 54.5  & 51.2 & 65.1 & 61.8 & 64.6 \\
\hdashline
Cultural + Diff. & 0.810 & 57.2 & \textbf{51.8} & \textbf{67.3} & \textbf{67.7} & \textbf{66.9} \\
\bottomrule
\end{tabular}%
}
\caption{\textbf{Downstream results across benchmarks.} Metrics are XCometXL (Flores), accuracy (G-MMLU, Include44, MGSM), and win-rate vs. Translated (mArenaHard, PolyWrite). Best values are bold.}
\label{tab:benchmarks}
\end{table*}

\subsubsection{Completion Quality}
Although the changes introduced in the prompts for the \textit{Naturalness} and \textit{Cultural adaptation} transformations are relatively small, the resulting completions differ substantially from those produced by the translated model (around 2$\times$ higher edit distance), as shown in \cref{tab:metrics_prompt_completion}, ``C'' columns. This suggests that even minor adjustments on the prompt side can lead to large shifts in completions. Notably, completions from the difficulty model are, on average, 2.2$\times$ longer than those from the translated model, i.e. yielding twice as many target-language tokens to train on. 
The effects of the individual transformations and the data mix overall correspond to the changes brought about in the prompt space.

We expected generations after the \textit{Naturalness} transformation to have a lower perplexity as a result of being more natural~\citep{li-etal-2025-lost}, but this is not indicated by the metric. One confounding factor might be that the perplexity scoring model is the pretrained model for our teacher model, which might bias the model towards prompts that it has altered more.
We next ask whether intervening on the prompts themselves induces greater naturalness in model responses.

\subsection{Downstream Performance}

\Cref{tab:benchmarks} summarizes the performance of the fine-tuned model across tasks, averaged across languages. We report a detailed language breakdown in \cref{app:breakdown}. In general, our transformations beat the translation-only baseline for all tasks and languages. We see surprisingly big differences in benchmark scores, given that we only exchanged max 10k prompts per language between variants.

\textbf{Beyond translationese.} 
We can see that the \textit{Naturalness} transformation, that is focused on increasing fluency and removing translation artifacts, brings only marginal gains on most benchmarks compared to the transformations that modify the content and domain of the prompts more.\footnote{\Cref{tab:comet_natural} shows that for some languages (cs, el, lt, eu, hu) this can be considered an improvement in translation quality, but the prompt is not strictly tied to post-editing.} This highlights the importance of going beyond translation: even if prompts were translated perfectly, their utility is limited by their content that is less relevant in other languages and cultures. Though, the naturalness transformation shines the most in open ended generation tasks: in mArenaHard it wins over the translated model by 7.7\% and even more in PolyWrite, which is focused on creative writing, winning by 13.8\%. 

\textbf{Cultural adaptation.} The gains in G-MMLU
and Include44 by 5.4\% (highest score overall) and 3.8\%, respectively, show that the cultural grounding of the prompts indeed helps for downstream knowledge retrieval in culturally relevant tasks. This reflects directly in the score of the cultural-sensitive subset of G-MMLU (\cref{tab:gmmlu_cultural_sensitive}), where this transformation provides a 7\% improvement, compared to 2\% for cultural-agnostic questions.

It also has beneficial effects on translation, math (+9.8\% accuracy wins over naturalized) and open-ended generation quality (e.g. +8\% win-rate on mArenaHard over naturalized prompts, especially high for Ukrainian and Slovak). Interestingly, the \textit{Difficulty} transformation also brings similar gains on Include44, which by closer inspection comes from questions in domains (see \cref{tab:include44_by_category}) centered around business, which likely have well-defined constraints and are more difficult in nature.

\textbf{The importance of difficulty.} 
The \textit{Difficulty} transformation, being most aggressive, also brings the overall largest benefits. It appears important for mathematical reasoning, as shown by the +8.6\% gains over only naturalized prompts. But more so in machine translation, achieving a notable improvement of +3.0 XCometXL points.\footnote{The 3.0 gain in XCometXL scores is estimated to be 95.3\% accurately aligned with humans~\citep{kocmi-etal-2024-navigating}.} 

\textbf{Combining complementary strengths.} \textit{Cultural adaptation} and \textit{Difficulty} transformations sometimes appear orthogonal in benefit to tasks like G-MMLU, mArenaHard, and PolyWrite. By mixing their data, taking 50\% each, we hope to achieve the best of both worlds. For MGSM and Include44, where they individually score similarly high, the gains add up to yield the best performance overall. For open-ended generation (mArenaHard and PolyWrite), the combined mix also scores highest, yielding an average win rate of 67.7\% and 66.9\% over translated prompts respectively. For the remaining tasks, the mix scores in between both, making this variant the overall most well-rounded. Future work may explore combinations through model merging rather than data mixing~\citep{aakanksha2024mixdatamergemodels}.

\begin{figure}[h!]
\centering
\begin{minipage}[c]{0.48\textwidth}\centering
\includegraphics[width=\linewidth]{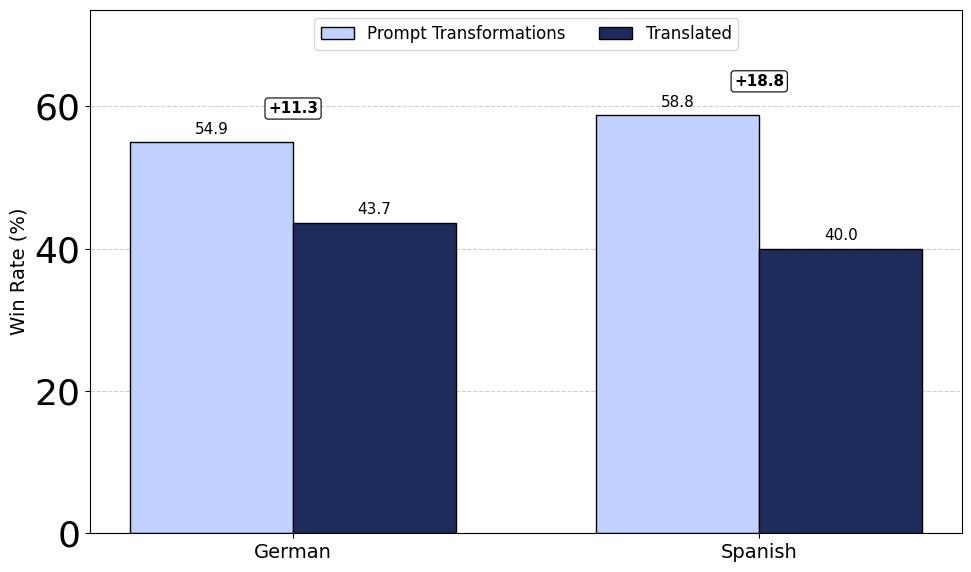}
\captionof{figure}{\textbf{Human evaluation on a PolyWrite subset.} Human preferences comparing \textit{Cultural+Difficulty Mix} with the \textit{Translated} baseline in two high-resource languages.}
\label{fig:human_eval_polywrite}
\end{minipage}\hfill
\begin{minipage}[c]{0.48\textwidth}\centering
\resizebox{\linewidth}{!}{\begin{tabular}{lcccc}
\toprule
 & \multicolumn{2}{c}{\textbf{Win-rate \%}} & \multicolumn{2}{c}{\textbf{Completion Length}} \\
\cmidrule(lr){2-3} \cmidrule(lr){4-5}
 & \textit{Ours} & Qwen & \textit{Ours} & Qwen \\
\midrule
mArenaHard & \textbf{56.8} & 43.2 & \textbf{5281} & 2548 \\
PolyWrite  & \textbf{88.4} & 11.6 & \textbf{3364} & 2208 \\
\bottomrule
\end{tabular}}
\captionof{table}{\textbf{Open-ended win rates vs. \textsc{Qwen2.5-7B}} with average completion lengths on mArenaHard and PolyWrite.}
\label{tab:open-ended-external}
\end{minipage}
\end{figure}

\textbf{Human evaluation.}
Automatic metrics and LLM judges are useful proxies, but they do not fully replace direct human preferences. We therefore evaluate a subset of PolyWrite prompts with human annotators in two high-resource languages, as these were the only settings where human annotation was feasible, and compare our best model (\textit{Cultural+Difficulty Mix}) against the \textit{Translated} baseline. \Cref{fig:human_eval_polywrite} shows a clear positive signal for our approach, even in high-resource languages where translation quality is already strong. Additionally, the annotators could optionally point out what type of issues had the completions: when our model's completion was preferred, among the most common reasons was because it was superior than the translated model response in terms of grammar, fluency and creativity (see \cref{tab:completion_issues}). This supports that prompt transformations improve generation quality beyond what automatic metrics alone can capture (setup details in \cref{app:human-eval}).

\needspace{22\baselineskip}
\subsection{Analysis}\label{sec:analysis}

\begin{figure}[h!]
\centering
\begin{minipage}[t]{0.58\textwidth}\centering
\includegraphics[width=\linewidth]{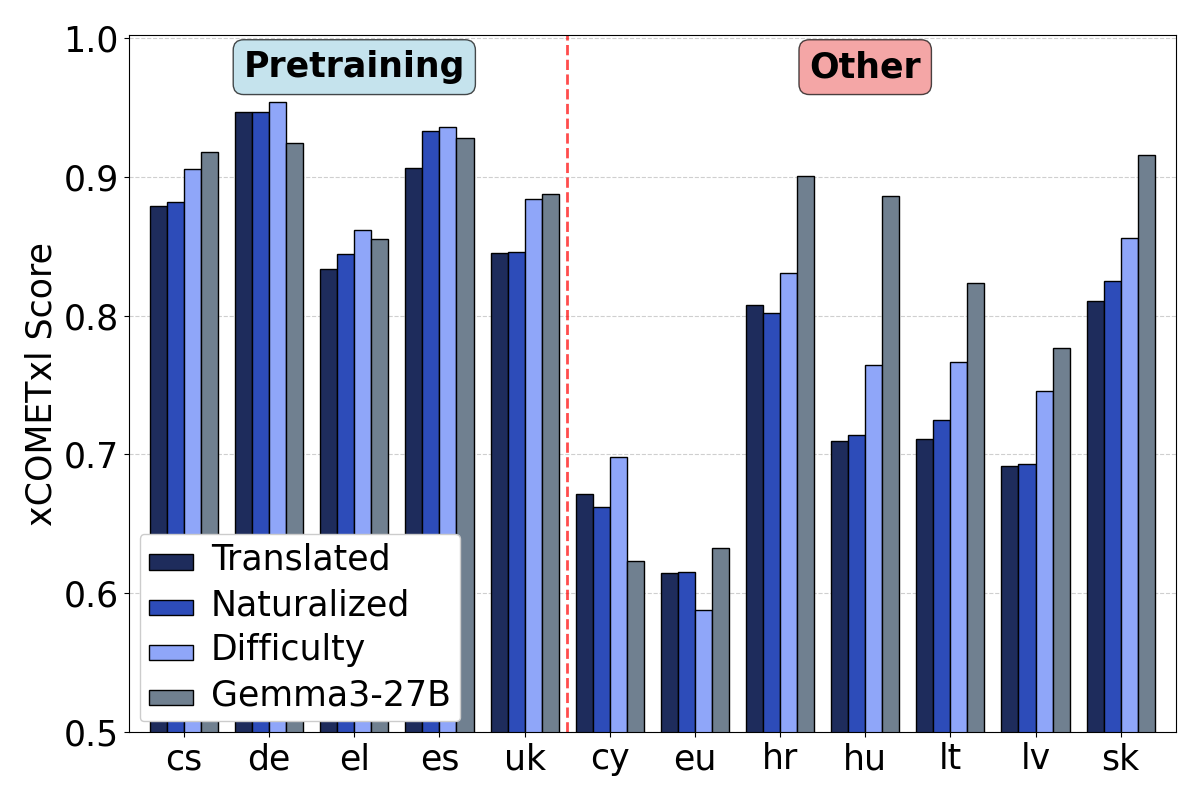}
\captionof{figure}{\textbf{Flores translation performance by language,} grouped by pretraining support, with teacher comparison.}
\label{fig:mt-against-teacher}
\end{minipage}\hfill
\begin{minipage}[t]{0.38\textwidth}\centering
\includegraphics[width=\linewidth]{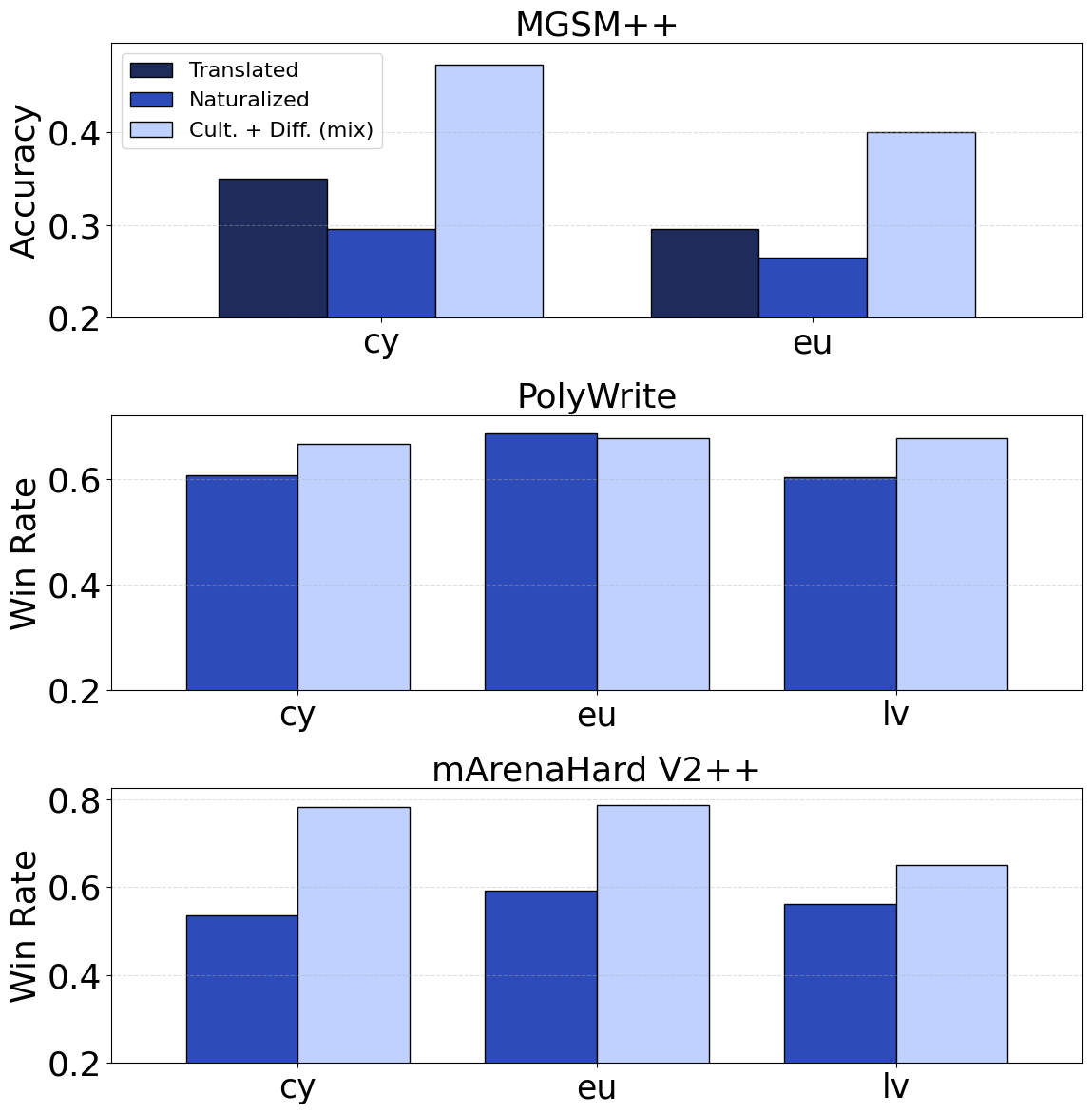}
\captionof{figure}{\textbf{Lowest-resource language performance} (Welsh, Basque, Latvian). Win rates are vs. \textit{Translated}.}
\label{fig:low_resource_languages}
\end{minipage}
\end{figure}

\subsubsection{What matters for quality?} In \cref{tab:quality_importance} (appendix) we break down multiple aspects of quality on the PolyWrite benchmark: diversity, naturalness and the ability to respond in the correct target language. We can see that our transformations improve over the translated variant in all aspects: downstream outputs are more natural, diverse and more likely in the right language. Similar to our prompt analysis we observe that diversity does not increase after \textit{Naturalness} transformation, but naturalness, as determined by an LLM judge, further increases. Due to our language id filtering, language confusion is rare across the bench but lowest in the mixed approach.

\subsubsection{How does language support and resourcedness affect performance?}
Naturally languages supported during pre-training show higher performance compared to those that were not supported (indicated in \cref{tab:languages}), e.g. on Flores the average \textit{Translated} baseline performance (\cref{fig:mt-against-teacher}) already diverges by 16.6 points in XCometXL between supported and unsupported languages (more details in \cref{tab:pretraining_support}).
However, our transformations significantly improve both groups relative to the baseline, the unsupported even more---by an average of +3.3 points (achieved by the \textit{Difficulty} model)---than the supported ones (+2.6 points on average).\footnote{\citet{kocmi-etal-2024-navigating} find this difference estimated to be 95.2\% accuracy with human preferences.} 
This is consistent with mArenaHard as well with +5.7 over \textit{Naturalized} for unsupported compared to +3.6 for supported) underlining the effectiveness of prompt optimization especially for cases of language expansion and under-served languages.

\subsubsection{Performance on Lowest-Resource Languages}

Our method depends on the performance of the translation model and teacher model. It is not well understood where the trade-off between noise and scale lie for synthetic data generation. In our study, there are very few cases where individual transformations did not yield downstream improvements over translations for individual languages. We particularly inspect the lowest-resource ones in \Cref{fig:low_resource_languages}: For MGSM, we find that for Basque and Welsh, the \textit{Naturalness} transformation performs worse than direct translation, but the other transformations succeed in improving over it, similar as for Welsh or Croatian in Flores (\cref{fig:mt-against-teacher}).
On the other hand, for mArenaHard, our overall best approach (\textit{Cultural+Difficulty}) yields substantial improvements over the light \textit{Naturalness} transformation for these languages. However, for PolyWrite, the data characteristics from this additional transformation does not provide the same benefit for win rates: it improves performance for Welsh and Latvian but does not give any gains over the \textit{Naturalness} transformation in Basque. Overall, these results highlight the nuanced relationship between translation quality, transformation strategy, and language resource level in determining the effectiveness of the prompt transformations.

\subsubsection{Comparison to External Models}
To ground our results in comparison with the external state of the art, we compare the performance against the teacher model itself, \textsc{Gemma3-27b-it}, focusing on generative performance in the machine translation task. 
The \textit{Difficulty} transformation performed particularly well---scoring within 2 points (and 3 points above the \textit{Translated} baseline) of the (more than 3 times larger) teacher on average. The breakdown by language in \cref{fig:mt-against-teacher} reveals that our method nominally outperforms the teacher model in German, Spanish, Greek, and Welsh, being most meaningful in Welsh and German and Spanish (96\%, 91\%, 70\% accuracy with humans respectively, according to~\citep{kocmi-etal-2024-navigating}).

Furthermore, we compare in \cref{tab:open-ended-external} the \textit{Cultural+Difficulty} model with \textsc{Qwen2.5-7B}\footnote{\url{https://huggingface.co/Qwen/Qwen2.5-7B-Instruct}} on open-ended generation tasks in both mArenaHard and PolyWrite. This evaluation aims to assess whether our interventions---though not explicitly optimized for downstream metrics on particular benchmarks---yield a competitive checkpoint. Our results show that the model achieves a 56\% win rate on mArenaHard on average across languages, winning in 9/13 languages, being outperformed in the highest-resourced ones, Czech, German, Spanish and English.
More impressively, our model achieves average win rates of 88\% on PolyWrite, winning in all languages but English, see individual language breakdowns in \cref{tab:winrates-qwen}. 
While for both evaluation sets, completion lengths for our model are substantially longer, it does not seem to be the deciding factor in win rates.
Upon inspection, we find that our model's generations for PolyWrite tend to be more elaborate, expressive and creative.

These outcomes highlight the effectiveness of our interventions in enhancing the model’s proficiency in generating text for our target languages.

\section{Related Work}

\subsection{Prompt Optimization Methods} Inference-time prompt optimization approaches iteratively rewrite instructions from task feedback, traces, or reflection signals, and improve performance for a fixed model~\citep{opsahlong2024miprov2,chen2024mapo,agrawal2025gepa, liu2024best,ajayi2026what}.
Our approach shares this intuition but shifts optimization to training time:
we rewrite prompts during data creation and target multilingual data deficiencies.

\subsection{Translation} \citet{enomoto-etal-2025-fair} compare instructing multilingual LLMs in the target languages with translation from English. They find that instructing models in English is not advantageous, when controlling for \textit{translationese}. To solve the \textit{translationese} problem, \citet{li-etal-2025-lost} proposes polishing or filtering unnatural data points after data generation in the instruction finetuning stage whereas we address it during the prompt expansion phase. The post-hoc filtering route was also chosen in Apertus~\citep{hernándezcano2025apertusdemocratizingopencompliant} and EuroLLM~\citep{martins2025eurollm9btechnicalreport}. \citet{martins2025eurollm9btechnicalreport} also explored synthetic prompt generation given LLM seed prompts from trusted sources.

\subsection{Naturalness} \citet{chen-etal-2024-good-data} showed that natural target-language data can outperform translated data for instruction tuning, particularly when evaluated on generative benchmarks and those written in target languages. Our work studies an attractive middle ground between the two scenarios: Even without native target language data (which are notoriously hard to get by), we can do much better than with plain translated prompts.

\subsection{Difficulty} \citet{xu2024wizardlm} demonstrated the effectiveness of using LLMs to synthetically scale-up the complexity of instructions based on the original prompts. \citet{muennighoff2025s1simpletesttimescaling} showed that even a small quantity of curated difficult mathematical problems can massively improve the performance.
Our work extends these findings to the multilingual setting and beyond math.

\subsection{Natural, Difficult, and Diverse} \citet{zhou2023lima} noted that only a handful of curated prompt-completion pairs from diverse and high-quality sources are needed to generate human-preferred responses.
Our work shows that through prompt transformations you can extend these curated English prompts to improve model performance in multilingual setting, addressing the massive imbalance of instruction data between languages.

\section{Conclusion}

In this work, we have demonstrated the potential of synthetic data generation for enriching multilingual datasets through the creation of higher-quality, contextually aligned, and culturally sensitive data that better reflects real-world language use. By systematically transforming the prompts, we are able to guide teacher model generations to be more adaptive and contextually nuanced to the target languages, consequently endowing our models with these characteristics. Our results, including automatic and human evaluation, show that systematic data transformations can produce models with outputs that are more natural, culturally grounded, creative, and linguistically rich. We position this study as an initial step toward principled approaches to multilingual synthetic data generation, an essential direction for developing inclusive, culturally aware, and globally capable language models.

\needspace{8\baselineskip}
\section*{Limitations}
\subsection{Reliability of Synthetic Data} Learning from synthetic data poses inherent risks and has well-studied limitations~\citep{liu2024best}. In our experiments, we found synthetic data beneficial and are not aware of similar human-authored data that could be used in its place.
We include language identification filtering, but other biases or errors could still have transferred from the teacher to the student model, especially those that would not show up in our evaluations. We recommend exploring the addition of more targeted filters in future work, especially for low-resource languages or languages where performance out-of-the-box is sub-par.
Ideally, native speakers should be involved to inspect samples of the generated data.

\subsection{Language Scope} Our study covers 12 languages, and we find consistent performance across the bench, taking into account their differences in resourcedness and support in student base and teacher model. Further work needs to confirm if the observations transfer to similarly resourced languages. We have not controlled for any benefits from geographic proximity. We have not pushed the method to the extremes (very low-resource and no evidence of language support) because it is obvious that relying on bootstrapping from existing models and automatic translation will not work with a cold start. However, there might be more languages beyond our lowest-resourced ones, that may benefit from our method.

\subsection{Evaluation Shortcomings} LLM judges might have been trained on \textit{translationese} as well and therefore favor it in their preferences~\citep{chen-etal-2024-good-data}. Similarly, the base model might still score \textit{translationese} as low-perplexity if it has seen translations during pre-training (likely according to \citet{thompson-etal-2024-shocking}).
Due to a lack of target-language evaluation benchmarks for challenging and relevant open-ended questions~\citep{kreutzer2025dj}, both our datasets for these tasks are composed of translated prompts for languages outside of English. Models trained on more translated prompts might have an advantage at evaluation time~\citep{chen-etal-2024-good-data,kreutzer2025dj}.

\subsection{Further Optimizing the Quality of the Data} Direction for further improvement of downstream results are the optimization of the machine translation, e.g. hand-picking the best available translator for each language and task, and the optimization of the generation process, e.g. by involving multiple teachers, quality filters or sequential edits~\citep{odumakinde-etal-2025-multilingual, khairi2025makingtakingbestn}.

\section*{Acknowledgments}
We thank our colleagues at Cohere Labs for their feedback throughout all stages of this project, including Thomas Euyang and Madeline Smith.

\bibliography{custom}

\newpage
\appendix

\section{Use of AI Disclosure}
For this paper we used AI to help with plotting code and grammar correction.

\section{Focus Languages}
\begin{table*}[h!]
\centering
\resizebox{1.0\textwidth}{!}{%
\begin{tabular}{lllcccccccccc}
\toprule
     \textbf{Language} & \textbf{Script} & \textbf{Lang. Family} &\multicolumn{2}{c}{\textbf{Resourcedness}} & \multicolumn{2}{c}{\textbf{WMT24++}} & \multicolumn{2}{c}{\textbf{Flores}} & \multicolumn{2}{c}{\textbf{Prompts}} \\
   (code) & & & Institution.  & Data &Expert & Gemma & Expert & Gemma & Expert & Gemma \\
     \midrule
      German (de)  & Latn & IE / Germanic & high&  5 & 91.91 & 83.64 & 97.78 & 92.41 & 93.96 & 92.49 \\
      Spanish (es) & Latn & IE / Italic & high&  5 &87.98 & 73.30 & 96.26 & 92.78 & 89.15 & 86.40 \\
      Czech (cs) & Latn & IE / Balto-Slavic & high  & 4 &86.11 & 69.37 & 96.64 & 91.91 & 86.00 & 82.03 \\
      Ukrainian (uk) & Cyrl & IE / Balto-Slavic & high   & 4 & 84.87 & 69.68 & 94.70 & 88.69 & 82.23 & 79.51 \\
      Greek (el) & Grek & IE / Greek & high &  3 & 84.62 & 69.86 & 93.37 & 85.58 & 83.86 & 80.78 \\
     $\star$Hungarian (hu)& Latn & Uralic / Finnic &high  & 4 &81.64 & 66.19 & 92.48 & 88.49 & 83.61& 78.88 \\
     $\star$Slovak (sk) & Latn & IE / Balto-Slavic & high &  3 &82.67 & 66.83 & 94.53 & 91.54 & 85.71 & 81.36 \\
     $\star$Croatian (hr)& Latn & IE / Balto-Slavic & high  & 3 &81.39 & 69.51 & 92.60 & 89.83 & 79.91 & 78.86 \\   
     $\star$Lithuanian (lt) & Latn & IE / Balto-Slavic &high & 3 &76.58 & 61.12 & 89.10 & 82.45 & 84.11 & 82.40 \\
     $\star$Latvian (lv) & Latn & IE / Balto-Slavic & high&  3 &64.15 & 60.90 & 76.86 & \textcolor{purple}{77.59}  & 69.65 & \textcolor{purple}{73.18} \\
     $\star$Basque (eu) & Latn & Basque & mid  & 4 & \multicolumn{2}{c}{---} & 62.87 & \textcolor{purple}{63.20} & 66.01 & \textcolor{purple}{70.19}\\
     $\star$Welsh (cy) & Latn & IE / Celtic
& mid &  3 &\multicolumn{2}{c}{---} & 79.22 & 62.02 &  73.75 & 68.30 \\    
     \midrule
     \textit{Avg} & -- &  -- &82.19 & 69.04 & 88.87 & 83.87 & 81.04 & 79.53\\
\bottomrule
\end{tabular}%
}
\caption{\textbf{Language Overview}: We characterize the languages of study (1) in terms of resourcedness with respect to data availability with levels estimated by \citet{ranathunga-de-silva-2022-languages} (\textit{Data}), and whether they are mid or high-institutional in terms of vitality according to Ethnologue (\textit{Institution.}) (data from \citep{ranathunga-de-silva-2022-languages}) 
and (2) in terms of XCometXL~\citep{guerreiro-etal-2024-xcomet} scores of the prompt translation model (\textit{Expert}, in-house expert model) and the transformation model (\textsc{Gemma3-27B-it}) on two traditional MT benchmarks: WMT24++ (en$\rightarrow \cdot$)~\citep{deutsch-etal-2025-wmt24} and Flores~\citep{nllb2022}, and a 1k sub-sample of our prompts, where we use XCometXL as a reference-free metric to estimate quality. Languages marked with $\star$ are not officially supported in the base model. IE stands for Indo-European.
}

\label{tab:languages}
\end{table*}

\Cref{tab:languages} compares translation quality on WMT24++~\citep{deutsch-etal-2025-wmt24} and Flores~\citep{nllb2022} and a subset of prompts, as measured by XCometXL~\citep{guerreiro-etal-2024-xcomet}. For prompt translation, we do not have references and use XCometXL as a quality estimation metric (i.e. call it without references).

\begin{table}[h!]
\centering
\resizebox{\textwidth}{!}{%
\begin{tabular}{lrrrrrrrrrrrr}
\toprule
\textbf{Dataset} & \textbf{cs} & \textbf{cy} & \textbf{de} & \textbf{el} & \textbf{es} & \textbf{eu} & \textbf{hr} & \textbf{hu} & \textbf{lt} & \textbf{lv} & \textbf{sk} & \textbf{uk} \\
\midrule
Translated              & 9992 & 9990 & 9998 & 9995 & 9990 & 9994 & 9987 & 9996 & 9993 & 9997 & 9990 & 9982 \\
\midrule
Naturalized             & 9583 & 9721 & 9570 & 9565 & 9516 & 9764 & 8740 & 9757 & 9752 & 9756 & 9597 & 9701 \\
Cultural                & 9449 & 9118 & 9417 & 8780 & 9355 & 9493 & 7760 & 9593 & 9557 & 9571 & 9240 & 9571 \\
Difficulty              & 9417 & 9276 & 9517 & 9396 & 9269 & 5287 & 8406 & 9662 & 9649 & 9599 & 9425 & 9603 \\
\hdashline
Cultural + Difficulty Mix & 9457 & 9218 & 9465 & 9127 & 9348 & 7439 & 8150 & 9639 & 9617 & 9607 & 9375 & 9608 \\
\hline
\end{tabular}%
}
\caption{\textbf{Number of samples per language}: Each sample consists of a prompt and its corresponding completion.}
\label{tab:language_counts}
\end{table}

\section{Transformation Prompts and Examples}
\label{app:examples}
\Cref{tab:prompts} lists the exact prompts we used for our transformations, and \cref{app-tab:examples-table} lists sample transformations.

\begin{table*}[]
    \centering
    \begin{tabular}{p{16cm}}
    \toprule
     \textbf{Transformation Prompts}\\
    \midrule
      \textit{Naturalness} \\
    \midrule
    You are an expert linguist and cultural adapter specializing in \{\texttt{language}\}. Rephrase the prompt to sound natural and authentic. Please adhere to the following guidelines in your naturalization:
    \begin{itemize}[nosep, label=-]
        \item Never answer instructions or questions, only rephrase them.
        \item Stay consistent in your rephrasing, always rephrase named entities in the same way.
        \item Match the formality level of the original text. In case of ambiguity, prefer an informal tone.
        \item Do not rephrase source code, but rephrase the comments within.
        \item When rephrasing JSON, your rephrasing must follow the exact same schema as the input. Do not rephrase JSON keys. Only rephrase the values.
        \item Follow the target language formatting conventions for dates and numbers.
    \end{itemize}    
    Here is the prompt to naturalize: \{\texttt{prompt}\}\\
    \midrule
    \midrule
    \textit{Cultural Adaptation} \\
    \midrule
    You are an expert cultural adapter for \{\texttt{language}\}. Adapt the prompt to be culturally appropriate and authentic. Follow these guidelines:
    \begin{itemize}[nosep, label=-]
        \item Do not answer the prompt; only adapt it culturally.
        \item Apply adaptation only when a reference would feel unnatural or out of place to a typical {language} speaker. Leave neutral/standardized items unchanged (technical instructions, standardized formats, URLs, file paths, measurements, ISO dates, library APIs). Do not fabricate facts or invent new culture-specific references.
        \item Preserve meaning, intent, and register. Keep personal names, trademarks, and official titles unless a well-established localized form exists.
        \item Locations: When a non-local place is incidental, swap it for a culturally equivalent local reference with similar connotation and social register. If the place is factual or essential to identity, keep it.
        \item Lexicon and orthography: Prefer native terms, spellings, and diacritics used in {language}. For common activities (e.g., sports), use the standard local term rather than anglicisms.
        \item Idioms and references: Replace idioms, holidays, and pop-culture references with well-known local equivalents; if none exists, paraphrase to preserve intent.
        \item Formatting: Use {language}-appropriate formats for dates, numbers, and units; convert units only when it aids comprehension without changing factual content.
        \item Code and structured data: Do not alter source code; comments may be adapted. For JSON, keep the same schema and keys; only adapt values.
    \end{itemize}    
    Here is the prompt to adapt: \{\texttt{prompt}\}\\
    \midrule
    \midrule
    \textit{Difficulty} \\
    \midrule
    You are an expert task designer for \{\texttt{language}\}. Rewrite the prompt to increase its difficulty and complexity while preserving the original intent and domain. Follow these guidelines:
    \begin{itemize}[nosep, label=-]
        \item Do not answer the prompt; only rewrite it to be more difficult.
        \item Preserve the original intent and domain. Do not change the topic or introduce factual errors.
        \item Tailor the rewrite to the prompt’s task type (e.g., coding, data wrangling/SQL, creative writing, analysis/explanation, classification, dialogue/roleplay).
        \item Code and JSON: do not alter source code; comments/instructions may be made more demanding. For JSON, keep the same schema and keys; only adapt values and constraints.
        \item Ensure the task remains solvable with the given information; if extra assumptions are needed, require the solver to list and justify them.
    \end{itemize}    
    Here is the prompt to complexify: \{\texttt{prompt}\}\\
         \bottomrule
    \end{tabular}
    \caption{Transformation prompt templates with placeholders for language and source prompt.}
    \label{tab:prompts}
\end{table*}
\input{tables/examples-table}

\section{Naturalness vs Translation}

Our naturalness transformation can be seen as a form of post-editing. Therefore, we evaluate the transformed prompts with a reference-free machine translation metric, XCometXL. \Cref{tab:comet_natural} shows that for most languages (7/12), translation quality is worsening after naturalization. This does not mean that quality decreases, it just means that it is less directly aligned with the source prompt before translation, according to the metric. 
Naturally, the languages with lowest translation quality have the highest potential for improving translation quality via the naturalness process, here Basque and Welsh.

\begin{table}[h!]
    \centering
    \begin{tabular}{lccc}
    \toprule
   	\textbf{Language} & \textbf{Translation }&	\textbf{Naturalized}	& \textbf{$\Delta$} \\
    \midrule
de  &	91.50	&91.11	&-0.39 \\
es  &	85.42	&84.88	& -0.56 \\
cs &	86.21&	83.27	&-2.95\\
hu 	&84.84	&85.25	&+0.41 \\
hr 	&80.45	&79.68	&-0.77\\
uk 	&85.34	&85.10	&-0.24 \\
el 	&82.46	&84.14&	+1.70\\
sk 	&80.21	&77.86 	&-2.35\\
lt 	&77.67	&79.07	&+1.40 \\
lv 	&68.07	&67.11	&-0.96 \\
eu &	62.53	&65.43	&+2.90 \\
cy	&69.62	&70.32&	+0.70\\
\midrule
\textit{Avg}	&\textit{79.53}	&\textit{79.43}	&\textit{-0.10}\\
    \bottomrule
    \end{tabular}
    \caption{\textbf{Translation Quality of Naturalized vs Machine-Translated}: We compare reference-free XCometXL scores before and after the naturalness transformation on a random sample of 100 prompts.}
    \label{tab:comet_natural}
\end{table}

\section{Training}\label{app:training}
We trained the model using supervised fine-tuning (SFT) with a cross-entropy loss function. Optimization was carried out using the Adam optimizer, configured with hyperparameters $\beta_{1}=0.9$, $\beta_{2}=0.95$, and $\epsilon=10^{-8}$. We applied an additive weight decay of $0.1$ and used gradient clipping with a maximum norm of $1.0$. Training was conducted for two epochs, and evaluation was performed using the final checkpoint. We used a batch size of 32 and employed a cosine learning rate decay schedule, with a peak learning rate of $2.5\times10^{-4}$ and an end learning rate of $1.25\times10^{-4}$.

\section{Evaluation}\label{app:evals}
\subsection{Overview}
\begin{table*}[h!]
    \centering
    \begin{tabular}{lll}
    \toprule
    \textbf{Name} & \textbf{Language List} & \textbf{Metric} \\
    \midrule
       GlobalMMLU  & cs, en, de, el, lt, es, uk & Accuracy\\
       Include44  & eu, hr, de, el, hu, lt, es, uk & Accuracy \\
       \midrule
       Flores & eu, hr, cs, de, el, hu, lv, lt, sk, es, uk, cy & xCometXL \\
       MGSM++ & eu, cs, en, de, el, hu, es, cy & Accuracy\\
       \midrule
       PolyWrite & en, de, es, cs, uk, hu, el, sk, hr, lt, lv, eu, cy & self-BLEU, win rates\\
       mArenaHard-v2.0++ & en, de, es, cs, uk, el, lv, lt, hu, hr, sk, cy, eu  & win rates\\
    \bottomrule
    \end{tabular}
    \caption{Evaluation suite: We evaluate on close-ended and open-ended tasks, which each cover a subset of our 13 focus languages. MGSM++ is an extension of MGSM~\citep{shi2023language} based on publicly available high-quality translations.
  }
    \label{tab:eval_suite}
    
\end{table*}
\Cref{tab:eval_suite} lists covered languages and metrics for each of the included benchmarks.

\subsection{Implementation}
For MGSM, use \texttt{simple-evals},\footnote{\url{https://github.com/openai/simple-evals}} for MCQA tasks we use the generative form of evaluation that operates on tokens and not on likelihoods, as also implemented in \texttt{simple-evals}. For MGSM++, the Basque translations were released in IberoBench ~\citep{baucells-etal-2025-iberobench},\footnote{\url{https://huggingface.co/datasets/HiTZ/MGSM-eu}} Greek curated by ILSP/Athena RC,\footnote{\url{https://huggingface.co/datasets/ilsp/mgsm_greek}} Welsh released by Language Technologies team from Bangor University,\footnote{\url{https://huggingface.co/datasets/techiaith/mgsm_cy}} Czech and Hungarian as curated for BenchMAX ~\citep{huang2025benchmaxcomprehensivemultilingualevaluation}.\footnote{\url{https://huggingface.co/datasets/LLaMAX/BenchMAX_Math}} For win-rates we use our own in-house implementation with LLM judge prompts specified in \cref{app:eval-prompts}. The order of models presented in preference ratings is shuffled.

\subsection{Human Evaluation}\label{app:human-eval}
For the human evaluation reported in \cref{fig:human_eval_polywrite}, we compare our best model (\textit{Cultural+Difficulty Mix}) against the \textit{Translated} baseline on a PolyWrite subset. We evaluate 80 prompts across two high-resource languages, and we use the same prompt set for both models to ensure a fair comparison. \Cref{tab:completion_issues} show the completions issues of the translated model when the \textit{Cultural+Difficulty Mix} model completion was preferred.

\begin{table*}[h!]
\centering
\begin{tabular}{lccc}
\hline
\textbf{Issue} & \textbf{Overall} & \textbf{German} & \textbf{Spanish} \\
\hline
Grammar & 23 & 13 & 10 \\
Fluency & 19 & 15 & 4 \\
Other & 17 & 8 & 9 \\
Creativity & 15 & 8 & 7 \\
Punctuation & 11 & 6 & 5 \\
Chattiness & 10 & 10 & 0 \\
Length & 9 & 8 & 1 \\
Naturalness & 9 & 7 & 2 \\
Repetition & 9 & 5 & 4 \\
Structure Flow & 6 & 5 & 1 \\
Tone & 4 & 2 & 2 \\
\hline
\end{tabular}
\caption{\textbf{Completion Quality Issues:} \textit{Translated} model completion issues when \textit{Cultural+Difficulty Mix} model completion was preferred.}
\label{tab:completion_issues}
\end{table*}

\begin{table}[h!]
\centering
\begin{tabular}{lccc}
\toprule
 & \textbf{Self-BLEU}$\downarrow$ & \textbf{NWR}$\uparrow$ & \textbf{LPR}$\uparrow$\\
\midrule
Translated & 33.73 & -- & 97.6\\
\midrule
Naturalized & 30.01 & 57.1 & 97.3\\
Cultural & 32.65 & 63.7 & 97.5\\
Difficulty & 34.01 & \textbf{69.3} & 97.3\\
\hdashline
Cultural + Diff. & \textbf{29.77} & 66.6 & \textbf{97.9}\\
\bottomrule
\end{tabular}
\caption{\textbf{PolyWrite quality metrics:} diversity (Self-BLEU), naturalness win-rate (NWR) vs. Translated, and language pass rate (LPR).}
\label{tab:quality_importance}
\end{table}

\section{Grouped results}
\subsection{Include44 Results by Domain}
\Cref{tab:include44_by_category} presents a domain breakdown of the Include44 results for each model.
\begin{table*}[h!]
\centering
\begin{tabular}{lccccc}
\toprule
\textbf{Category} & \textbf{Business} & \textbf{Culture} & \textbf{Health} & \textbf{Other} & \textbf{STEM} \\
\midrule
Translated          & 0.524 & 0.524 & 0.320 & 0.396 & 0.410 \\
\midrule
Naturalized         & 0.627 & 0.515 & 0.372 & 0.372 & 0.439 \\
Cultural            & 0.534 & 0.568 & 0.372 & 0.399 & \textbf{0.482} \\
Difficulty          & \textbf{0.689} & 0.581 & \textbf{0.399} & \textbf{0.372} & 0.444 \\
\hdashline
Cult. + Diff. (mix) & 0.588 & \textbf{0.587} & 0.393 & 0.411 & 0.447 \\
\bottomrule
\end{tabular}
\caption{\textbf{Include44 Accuracy By Domain}: average accuracy for each model by question domain. Highest value per domain is highlighted in bold.}
\label{tab:include44_by_category}
\end{table*}

\subsection{G-MMLU Cultural Sensitivity}
\Cref{tab:gmmlu_cultural_sensitive} presents Global-MMLU results for questions annotated as cultural-agnostic(CA) and cultural-sensitive(CS). Each category contains 200 questions and annotations are only available for English, German and Spanish.
\begin{table*}[h!]
\centering
\begin{tabular}{lcc}
\toprule
\textbf{Model} & \textbf{Cultural-Agnostic (CA)} & \textbf{Cultural-Sensitive (CS)} \\
\midrule
Translated                & 0.677 & 0.605 \\
\midrule
Naturalized               & 0.676 & 0.635 \\
Cultural                  & 0.700 & \textbf{0.670} \\
Difficulty                & \textbf{0.708} & 0.666 \\
\hdashline
Cultural + Difficulty Mix & 0.683 & 0.642 \\
\bottomrule
\end{tabular}
\caption{\textbf{Performance on GMMLU by cultural sensitivity.} Comparison of model accuracies on cultural-agnostic (CA) and cultural-sensitive (CS) subsets. Higher is better. Only includes English, German and Spanish, the CS-annotated languages that overlap with our subset of languages. }
\label{tab:gmmlu_cultural_sensitive}
\end{table*}

\subsection{Supported vs Unsupported performance}
\Cref{tab:pretraining_support} presents performance comparison on Flores for languages grouped by supported languages (languages that were included during pretraining of the base model) and unsupported languages. List of supported languages is showed in \cref{model}.
\begin{table}[h!]
\centering
\begin{tabular}{lcc}
\hline
\textbf{Model} & \textbf{Unsupported} & \textbf{Supported} \\
\hline
Translated          & 0.716& 0.882\\
\midrule
Naturalized         & 0.719& 0.890\\
Cultural            & 0.732& 0.907\\
Difficulty          & 0.749& 0.908\\
\hdashline
Cultural + Difficulty Mix & 0.742& 0.905\\
\hline
\end{tabular}
\caption{Performance comparison in machine translation (XCometXL score on Flores), averaged across languages grouped by pretraining support (see \cref{tab:languages}).}
\label{tab:pretraining_support}
\end{table}

\subsection{Downstream Results By Language}\label{app:breakdown}
\Cref{tab:marena}, \cref{tab:polywrite}, \cref{tab:globalmmlu}, \cref{tab:include44}, \cref{tab:flores} and \cref{tab:mgsm} present language breakdowns for the performance of each model on the benchmarks. \Cref{tab:winrates-qwen} presents the language breakdowns for the performance of the \textit{Cultural+Difficulty} model against an external model in open-ended generation.

\begin{table*}[h!]
\centering
\resizebox{\textwidth}{!}{
\begin{tabular}{lcccccccccccccc}
\toprule
\textbf{Model} & \textbf{cs} & \textbf{cy} & \textbf{de} & \textbf{el} & \textbf{en} & \textbf{es} & \textbf{eu} & \textbf{hr} & \textbf{hu} & \textbf{lt} & \textbf{lv} & \textbf{sk} & \textbf{uk} & \textbf{Average} \\
\midrule
Naturalized          & 0.552 & 0.537* & 0.533* & 0.550 & 0.800 & 0.568 & 0.592 & 0.517* & 0.545 & 0.600 & 0.562 & 0.558 & 0.600 & 0.578 \\
Cultural             & \textbf{0.623} & 0.606 & 0.646 & 0.629 & 0.846 & \textbf{0.700} & 0.622 & 0.636 & 0.640 & 0.666 & 0.602 & 0.601 & 0.723 & 0.657 \\
Difficulty           & 0.597 & 0.564 & \textbf{0.653} & \textbf{0.630} & 0.815 & 0.667 & 0.544* & 0.578 & 0.605 & 0.587 & 0.655 & 0.540* & 0.600 & 0.618 \\
\hdashline
Cult. + Diff. Mix & 0.617 & \textbf{0.685} & 0.630 & 0.643 & \textbf{0.851} & 0.692 & \textbf{0.623} & \textbf{0.676} & \textbf{0.714} & \textbf{0.683} & \textbf{0.663} & \textbf{0.654} & \textbf{0.662} & \textbf{0.676} \\
\bottomrule
\end{tabular}%
}
\caption{\textbf{mArenaHard++ results:} win-rates over translated baseline. Best score per language in bold. Values marked with an asterisks indicate win-rate differences are not significant according to 95\% CIs.}
\label{tab:marena}
\end{table*}

\begin{table*}[h!]
\centering
\resizebox{\textwidth}{!}{
\begin{tabular}{lcccccccccccccc}
\toprule
\textbf{Model} & \textbf{cs} & \textbf{cy} & \textbf{de} & \textbf{el} & \textbf{en} & \textbf{es} & \textbf{eu} & \textbf{hr} & \textbf{hu} & \textbf{lt} & \textbf{lv} & \textbf{sk} & \textbf{uk} & \textbf{Average} \\
\midrule
Naturalized                 & 0.682 & 0.607 & 0.662 & \textbf{0.626} & 0.632 & 0.684 & 0.686 & 0.597 & 0.623 & 0.600 & 0.603 & \textbf{0.675} & 0.617 & 0.638 \\
Cultural                    & \textbf{0.692} & 0.613 & 0.701 & 0.619 & 0.594 & 0.774 & 0.677 & \textbf{0.649} & 0.672 & 0.665 & 0.626 & 0.630 & \textbf{0.682} & 0.661\\
Difficulty                  & 0.646 & 0.598 & \textbf{0.766} & 0.555* & 0.574* & \textbf{0.787} & \textbf{0.697} & 0.571* & \textbf{0.695} & \textbf{0.710} & 0.594 & 0.552* & 0.656 & 0.646 \\
\hdashline
Cultural + Difficulty Mix   & 0.591 & \textbf{0.668} & 0.740 & 0.568* & \textbf{0.742} & 0.755 & 0.677 & \textbf{0.649} & 0.682 & 0.684 & \textbf{0.677} & 0.623 & 0.643 & \textbf{0.669} \\

\bottomrule
\end{tabular}%
}
\caption{\textbf{PolyWrite results}: win-rates over Translated baseline. Best score per language in bold. Values marked with an asterisks indicate win-rate differences are not significant according to 95\% CIs.}
\label{tab:polywrite}
\end{table*}

\begin{table*}[h!]
\centering
\begin{tabular}{lcccccccc}
\toprule
\textbf{Model} & \textbf{cs} & \textbf{de} & \textbf{el} & \textbf{en} & \textbf{es} & \textbf{lt} & \textbf{uk} & \textbf{Average} \\
\midrule
Translated                  & 0.529 & 0.533 & 0.506 & 0.605 & 0.558 & 0.440 & 0.503 & 0.525\\
\midrule
Naturalized                 & 0.535 & 0.546 & 0.507 & 0.602 & 0.567 & 0.450 & 0.510 & 0.531\\ 
Cultural                    &  \textbf{0.579} & \textbf{0.590} & \textbf{0.556} & \textbf{0.665} & \textbf{0.619} & 0.486 & 0.554 & \textbf{0.579}\\
Difficulty                  & 0.555 & 0.535 & 0.540 & 0.575 & 0.565 & \textbf{0.495} & \textbf{0.555} & 0.545 \\
\hdashline
Cultural + Difficulty Mix   & 0.573 & 0.583 & 0.550 & 0.649 & 0.613 & 0.488 & 0.551 & 0.572 \\
\bottomrule
\end{tabular}
\caption{\textbf{Global MMLU results}: Accuracy. Best score per language in bold.}
\label{tab:globalmmlu}
\end{table*}

\begin{table*}[h!]

\resizebox{\textwidth}{!}{
\centering
\begin{tabular}{lccccccccc}
\toprule
\textbf{Model} & \textbf{de} & \textbf{el} & \textbf{es} & \textbf{eu} & \textbf{hr} & \textbf{hu} & \textbf{lt} & \textbf{uk} & \textbf{Average} \\
\midrule
Translated                  & 0.489 & 0.409 & 0.544 & 0.358 & 0.556 & 0.373 & 0.489 & 0.542 & 0.470 \\
\midrule
Naturalized                 & 0.460 & 0.427 & 0.575 & 0.334 & 0.567 & 0.391 & 0.453 & 0.542 & 0.469 \\
Cultural                    & 0.496 & 0.447 & 0.613 & 0.346 & 0.620 & 0.405 & 0.530 & \textbf{0.609} & 0.508\\
Difficulty                  & \textbf{0.518} & 0.465 & 0.598 & 0.336 & \textbf{0.651} & \textbf{0.407} & \textbf{0.539} & 0.580 & 0.512 \\
\hdashline
Cultural + Difficulty Mix   &  0.482 & \textbf{0.480} & \textbf{0.644} & \textbf{0.372} & 0.644 & 0.405 & 0.530 & 0.584 & \textbf{0.518}\\

\bottomrule
\end{tabular}}
\caption{\textbf{Include 44 results}: Accuracy. Best score per language in bold.}
\label{tab:include44}
\end{table*}

\begin{table*}[h!]
\centering
\resizebox{\textwidth}{!}{
\begin{tabular}{lccccccccccccc}
\toprule
\textbf{Model} & \textbf{cs} & \textbf{cy} & \textbf{de} & \textbf{el} & \textbf{es} & \textbf{eu} & \textbf{hr} & \textbf{hu} & \textbf{lt} & \textbf{lv} & \textbf{sk} & \textbf{uk} & \textbf{Average} \\
\midrule
Translated          & 0.879 & 0.671 & 0.947 & 0.834 & 0.906 & 0.614 & 0.808 & 0.710 & 0.711 & 0.692 & 0.810 & 0.845 & 0.786 \\
\midrule
Naturalized         & 0.882 & 0.662 & 0.946 & 0.844 & 0.933 & 0.615 & 0.802 & 0.714 & 0.725 & 0.693 & 0.825 & 0.846 & 0.791 \\
Cultural            & 0.904 & 0.673 & \textbf{0.954} & 0.858 & 0.935 & 0.614 & 0.821 & 0.733 & 0.736 & 0.707 & 0.842 & 0.881 & 0.805 \\
Difficulty          & 0.905 & \textbf{0.698} & \textbf{0.954} & \textbf{0.861} & \textbf{0.936} & 0.588 & 0.831 & 0.764 & 0.767 & 0.746 & 0.856 & 0.884 & 0.816 \\
Cultural + Difficulty Mix & 0.900 & 0.684 & \textbf{0.954} & 0.856 & 0.933 & 0.601 & 0.822 & 0.747 & 0.757 & 0.731 & 0.855 & 0.880 & 0.810 \\
\midrule
\textsc{Gemma3-27B} & \textbf{0.918} & 0.623 & 0.924 & 0.855 & 0.928 & \textbf{0.633} & \textbf{0.901} & \textbf{0.886} & \textbf{0.824} & \textbf{0.776} & \textbf{0.916} & \textbf{0.888} & \textbf{0.839} \\
\bottomrule
\end{tabular}}
\caption{\textbf{Flores}: XCometXL scores. Best score per language in bold.}
\label{tab:flores}
\end{table*}

\begin{table*}[h!]
\centering
\resizebox{\textwidth}{!}{
\begin{tabular}{lccccccccc}
\toprule
\textbf{Model} & \textbf{cs} & \textbf{cy} & \textbf{de} & \textbf{el} & \textbf{en} & \textbf{es} & \textbf{eu} & \textbf{hu} & \textbf{Average} \\
\midrule
Translated                  & 0.665 & 0.350 & 0.645 & 0.625 & 0.745 & 0.705 & 0.295 & 0.470 & 0.5625 \\
\midrule
Naturalized                 & 0.640 & 0.295 & 0.665 & 0.680 & 0.750 & 0.735 & 0.265 & 0.490 & 0.5650 \\
Cultural                    & 0.736 & 0.456 & 0.740 & 0.752 & 0.840 & \textbf{0.800} & 0.376 & 0.580 & 0.660 \\
Difficulty                  & 0.750 & 0.385 & \textbf{0.775} & \textbf{0.780} & 0.810 & 0.790 & 0.320 & \textbf{0.595} & 0.6506 \\
\hdashline
Cultural + Difficulty Mix   & \textbf{0.780} & \textbf{0.472} & 0.756 & 0.752 & \textbf{0.876} & 0.788 & \textbf{0.400} & 0.556 & \textbf{0.673} \\
\bottomrule
\end{tabular}}
\caption{\textbf{MGSM++}: best score per language in bold.}
\label{tab:mgsm}
\end{table*}

\begin{table*}[h!]
\centering
\resizebox{\textwidth}{!}{
\begin{tabular}{lcccccccccccccc}
\toprule
\textbf{Benchmark} & \textbf{cs} & \textbf{cy} & \textbf{de} & \textbf{el} & \textbf{en} & \textbf{es} & \textbf{eu} & \textbf{hr} & \textbf{hu} & \textbf{lt} & \textbf{lv} & \textbf{sk} & \textbf{uk} & \textbf{Average} \\
\midrule
mArenaHard   & 0.427 & \textbf{0.794} & 0.438 & \textbf{0.696} & 0.226 & 0.421 & \textbf{0.766} & \textbf{0.578} & \textbf{0.603} & \textbf{0.660} & \textbf{0.648} & \textbf{0.568} & \textbf{0.556} & \textbf{0.568} \\
PolyWrite  & \textbf{0.942} & \textbf{0.794} & \textbf{0.890} & \textbf{0.994} & 0.284 & \textbf{0.768} & \textbf{0.974} & \textbf{0.974} & \textbf{0.987} & \textbf{0.981} & \textbf{0.948} & \textbf{0.987} & \textbf{0.974} & \textbf{0.884} \\

\bottomrule
\end{tabular}%
}
\caption{\textbf{Open Ended Win-Rates against Qwen2.5-7B}: Win-rates across languages from direct comparisons of the \textit{Cultural+Difficulty} model against Qwen2.5-7B on mArenaHard and
PolyWrite. Language where our model wins are in bold. All win-rate differences are significant according to 95\% CIs.}
\label{tab:winrates-qwen}
\end{table*}

\section{Evaluation Prompts}\label{app:eval-prompts}
For general win-rates we follow the LLM judge prompt proposed in~\citep{ustun-etal-2024-aya} that not only asks the judge to select the more correct generation, but also specifies the target language and indicates that it should be grammatically correct and fluent. For naturalness win-rates we use the prompt defined in \cref{tab:naturalness_prompt}.
To grade prompts and completions according to difficulty and quality using an LLM as described in \cref{data_evals}, we used the prompt defined in \cref{tab:quality_difficulty_prompt}.

\begin{table*}[]
    \centering
    \begin{tabular}{p{15cm}}
    \toprule
     \textbf{Completion Grader (Quality \& Difficulty Evaluation)} \\
    \midrule
    You are a strict grader of responses. The questions are enclosed in \texttt{<question></question>} tags, and the answers are enclosed in \texttt{<answer></answer>} tags. Given a question–answer pair, evaluate the \texttt{<answer>} according to the following criteria: \\
    \begin{itemize}[nosep, label=-]
        \item Does the \texttt{<answer>} address the \texttt{<question>} fully in all parts?
        \item Is the \texttt{<question>} logically sound?
        \item Is the \texttt{<answer>} logically sound?
        \item Is the \texttt{<answer>} factually correct and coherent?
        \item Does the \texttt{<answer>} contain any hallucinations?
        \item Is the \texttt{<answer>} properly formatted?
        \item Does the \texttt{<answer>} use correct punctuation for the given language?
        \item Are there any other issues with the \texttt{<question>} or \texttt{<answer>} not covered above?
    \end{itemize} \\\\
    Any \texttt{<question>} or \texttt{<answer>} that fails to meet the above criteria should be penalized accordingly. \\
    \begin{itemize}[nosep, label=-]
        \item  Give the question and answer pair a quality rating between [A,B,C,D,E] with A being best and E being worst. The rating should be at the very end and inside the tag "<quality> </quality>" without markdown formatting.
        \item Give the question and answer pair a difficulty rating between [easy, medium, hard]. The rating should come after the quality rating and inside the tag "<difficulty> </difficulty>" without markdown formatting.
    \end{itemize}\\
    \midrule 
    \midrule
    \textbf{Prompt Grader (Prompt Quality \& Difficulty Evaluation)} \\
    \midrule
    You are a strict Grader of prompts. The prompt is inside \texttt{<prompt></prompt>} tags. Given the following \texttt{<prompt>}, please grade it based on the following criteria: \\
    \begin{itemize}[nosep, label=-]
        \item Is the \texttt{<prompt>} clear and unambiguous?
        \item Is the \texttt{<prompt>} logically sound and coherent?
        \item Is the \texttt{<prompt>} factually correct?
        \item Does the \texttt{<prompt>} provide sufficient context for a meaningful response?
        \item Is the \texttt{<prompt>} well-structured and grammatically correct?
        \item Does the \texttt{<prompt>} have a clear objective or question?
        \item Is the \texttt{<prompt>} appropriately scoped (not too broad or too narrow)?
        \item Does the \texttt{<prompt>} avoid contradictions or false premises?
    \end{itemize} \\\\
    Any \texttt{<prompt>} that fails to address any of the criteria sufficiently should be penalized accordingly. \\\\
    Please think step-by-step and elaborate first before doing the following: \\
    \begin{itemize}[nosep, label=-]
        \item Give the prompt a quality rating between [A,B,C,D,E] with A being best and E being worst. The rating should be at the very end and inside the tag "<quality> </quality>" without markdown formatting.
        \item Give the prompt a difficulty rating between [easy, medium, hard]. The rating should come after the quality rating and inside the tag "<difficulty> </difficulty>" without markdown formatting.
    \end{itemize}
    \\
    \bottomrule
    \end{tabular}
    \caption{Grader prompts to evaluate completions and prompts for overall quality and difficulty.}
    \label{tab:quality_difficulty_prompt}
\end{table*}

\begin{table*}[]
    \centering
    \begin{tabular}{p{15cm}}
    \toprule
     \textbf{Judge Prompt }\\
    \midrule
    Which of the following responses is the \textbf{most natural-sounding and overall best} one for the given instruction in \{\texttt{language}\}? A good response should follow these rules, \textbf{with a primary focus on Rule 5 (Naturalness)}:
    \begin{itemize}[nosep, label=-]
        \item 1) It should be in \{\texttt{language}\}.
        \item 2) It should complete the request in the instruction.
        \item 3) It should be factually correct and semantically comprehensible.
        \item 4) It should be grammatically correct and fluent.
        \item 5) \textbf{Crucially, it should sound natural in \{\texttt{language}\}.} This means it uses common phrasing, appropriate tone, and idiomatic expressions (where suitable) that a native speaker would typically use. It should avoid awkward, stilted, or "translated-sounding" sentences.
\end{itemize}\\\\
Instruction: \{\texttt{prompt}\}

Response (A): \{\texttt{completion\_a}\}

Response (B): \{\texttt{completion\_b}\}\\\\

FIRST provide a concise comparison of the two responses, \textbf{evaluating primarily which response sounds more natural and authentic in \{\texttt{language}\}}. Consider if it uses common phrasing and tone typical of a native speaker, while also meeting the other criteria (completeness, correctness, grammar). If one Response is better, explain which you prefer and why, highlighting differences in naturalness. If both responses are identical or equally good or bad (especially in terms of naturalness), explain why.

SECOND, on a new line, state exactly one of 'Response (A)' or 'Response (B)' or 'TIE' to indicate your choice of preferred response.\\\\

Your response should use the format:

Comparison: <concise comparison and explanation, focusing on naturalness>

Preferred: <'Response (A)' or 'Response (B)' or 'TIE'>\\

         \bottomrule
    \end{tabular}
    \caption{Judge prompt to do pairwise open-ended evaluation focused on naturalness.}
    \label{tab:naturalness_prompt}
\end{table*}

\end{document}

%% file: tables/examples-table.tex
\begin{table*}
\centering
\small
\begin{tabular}{p{2cm}p{14.5cm}}
\toprule
Transformation & Result \\
\midrule

Seed prompt & \begin{minipage}[t]{\linewidth}\footnotesize\ttfamily  What's the difference between thai ginger and regular ginger?\end{minipage} \\
\addlinespace
Translation(de) & \begin{minipage}[t]{\linewidth}\footnotesize\ttfamily  Was ist der Unterschied zwischen thailändischem Ingwer und normalem Ingwer?\end{minipage} \\
\addlinespace
Naturalisation & \begin{minipage}[t]{\linewidth}\footnotesize\ttfamily Was ist der Unterschied zwischen thailändischem Ingwer und dem Ingwer, den man normalerweise kennt?\end{minipage}  \\
\addlinespace
Cultural Adaptation & \begin{minipage}[t]{\linewidth}\footnotesize\ttfamily Was ist der Unterschied zwischen thailändischem Ingwer und dem Ingwer, den man hierzulande üblicherweise im Handel findet?\end{minipage} \\
\addlinespace
Difficulty & \begin{minipage}[t]{\linewidth}\footnotesize\ttfamily Vergleichen und kontrastieren Sie detailliert thailändischen Ingwer (Kha) mit dem in der europäischen Küche üblicherweise verwendeten Ingwer (Zingiber officinale). Ihre Analyse soll folgende Aspekte berücksichtigen: botanische Unterschiede (einschließlich, aber nicht beschränkt auf Rhizom-Morphologie und ätherische Öl-Zusammensetzung), typische Anbaubedingungen und -methoden in Thailand versus anderen Anbauregionen, kulinarische Anwendungen und Geschmacksnuancen (unter Berücksichtigung regionaler Variationen innerhalb Thailands und Europas), sowie traditionelle medizinische Verwendungen in den jeweiligen Kulturen.\textbackslash n\textbackslash nIhre Antwort soll in Form eines strukturierten Essays verfasst sein, der mindestens 500 Wörter umfasst.  Zitieren Sie mindestens drei wissenschaftliche Quellen (z.B. botanische Datenbanken, ethnobotanische Studien, landwirtschaftliche Forschungsberichte) und zwei Kochbücher oder kulinarische Fachartikel, die sich explizit mit Ingwer befassen.\textbackslash n\textbackslash nDarüber hinaus sollen Sie eine Tabelle erstellen, die die wichtigsten Unterschiede in den oben genannten Aspekten übersichtlich zusammenfasst.\textbackslash n\textbackslash nSollten Sie Annahmen über spezifische Ingwersorten innerhalb Europas treffen (z.B. hinsichtlich Herkunft oder Anbaumethoden), so sind diese explizit zu benennen und zu begründen.  Ebenso sind etwaige Schwierigkeiten bei der Beschaffung spezifischer Informationen zu Kha zu erwähnen und die angewandten Strategien zur Informationsfindung zu erläutern.\end{minipage} \\

\addlinespace
\hline
\addlinespace

Seed prompt & \begin{minipage}[t]{\linewidth}\footnotesize\ttfamily Is it possible to have a permanent job and a part-time job at the same time in the UK?\end{minipage} \\
\addlinespace
Translation (es) & \begin{minipage}[t]{\linewidth}\footnotesize\ttfamily ¿Es posible tener un trabajo permanente y un trabajo a tiempo parcial al mismo tiempo en el Reino Unido?\end{minipage} \\
\addlinespace
Naturalisation & \begin{minipage}[t]{\linewidth}\footnotesize\ttfamily ¿Se puede tener un trabajo fijo y otro a media jornada a la vez en el Reino Unido? \end{minipage}\\
\addlinespace
Cultural Adaptation & \begin{minipage}[t]{\linewidth}\footnotesize\ttfamily ¿Es posible compaginar un trabajo a tiempo completo con otro a media jornada en el Reino Unido?\end{minipage} \\
\addlinespace
Difficulty & \begin{minipage}[t]{\linewidth}\footnotesize\ttfamily Un ciudadano español está considerando una oferta de empleo a tiempo parcial en el Reino Unido mientras ya mantiene un contrato de trabajo a tiempo completo con una empresa española que permite el trabajo remoto desde el extranjero. Analice exhaustivamente la legalidad y las implicaciones fiscales de esta situación, considerando tanto la legislación laboral del Reino Unido como las regulaciones fiscales españolas e internacionales (incluyendo posibles convenios de doble imposición). Su respuesta debe:\textbackslash n\textbackslash n1.  **Detallar las restricciones legales del Reino Unido** con respecto a la compatibilidad de empleos, incluyendo la consideración de las cláusulas de exclusividad en los contratos de trabajo (si las hubiera) y las regulaciones específicas para trabajadores extranjeros.\textbackslash n2.  **Explicar las obligaciones fiscales del ciudadano español** en España, incluyendo cómo se tratarían los ingresos obtenidos en el Reino Unido y si estaría sujeto a algún tipo de declaración especial.\textbackslash n3.  **Evaluar las posibles implicaciones en la Seguridad Social** tanto en España como en el Reino Unido, considerando la posibilidad de cotizaciones en ambos países y cómo esto afectaría a sus derechos a futuras prestaciones.\textbackslash n4.  **Identificar y justificar cualquier suposición** que deba hacerse para completar el análisis, por ejemplo, sobre el tipo de contrato de trabajo en España, el nivel de ingresos en ambos empleos, o la existencia de un convenio de doble imposición aplicable.\textbackslash n5.  **Presentar un resumen conciso** de los riesgos y beneficios clave para el ciudadano español, incluyendo recomendaciones sobre los pasos a seguir para garantizar el cumplimiento legal y fiscal.\textbackslash n\textbackslash nLa respuesta debe estar redactada en un español formal y preciso, demostrando un conocimiento profundo de la legislación laboral y fiscal relevante. Se valorará la capacidad de presentar información compleja de manera clara y organizada. \end{minipage} \\
\addlinespace
\bottomrule
\end{tabular}
\caption{Example prompts and their respective transformations for German (de) above and Spanish (es) below.}
\label{app-tab:examples-table}
\end{table*}

%% file: custom.bib
@misc{he2024doespromptformattingimpact,
      title={Does Prompt Formatting Have Any Impact on LLM Performance?}, 
      author={Jia He and Mukund Rungta and David Koleczek and Arshdeep Sekhon and Franklin X Wang and Sadid Hasan},
      year={2024},
      eprint={2411.10541},
      archivePrefix={arXiv},
      primaryClass={cs.CL},
      url={https://arxiv.org/abs/2411.10541}, 
}

@misc{aakanksha2024mixdatamergemodels,
      title={Mix Data or Merge Models? Optimizing for Diverse Multi-Task Learning}, 
      author={Aakanksha and Arash Ahmadian and Seraphina Goldfarb-Tarrant and Beyza Ermis and Marzieh Fadaee and Sara Hooker},
      year={2024},
      eprint={2410.10801},
      archivePrefix={arXiv},
      primaryClass={cs.CL},
      url={https://arxiv.org/abs/2410.10801}, 
}

@inproceedings{kocmi-etal-2024-navigating,
    title = "Navigating the Metrics Maze: Reconciling Score Magnitudes and Accuracies",
    author = "Kocmi, Tom  and
      Zouhar, Vil{\'e}m  and
      Federmann, Christian  and
      Post, Matt",
    editor = "Ku, Lun-Wei  and
      Martins, Andre  and
      Srikumar, Vivek",
    booktitle = "Proceedings of the 62nd Annual Meeting of the Association for Computational Linguistics (Volume 1: Long Papers)",
    month = aug,
    year = "2024",
    address = "Bangkok, Thailand",
    publisher = "Association for Computational Linguistics",
    url = "https://aclanthology.org/2024.acl-long.110/",
    doi = "10.18653/v1/2024.acl-long.110",
    pages = "1999--2014"
}

@inproceedings{bizzoni-lapshinova-koltunski-2021-measuring,
    title = "Measuring Translationese across Levels of Expertise: Are Professionals more Surprising than Students?",
    author = "Bizzoni, Yuri  and
      Lapshinova-Koltunski, Ekaterina",
    editor = "Dobnik, Simon  and
      {\O}vrelid, Lilja",
    booktitle = "Proceedings of the 23rd Nordic Conference on Computational Linguistics (NoDaLiDa)",
    month = may # " 31--2 " # jun,
    year = "2021",
    address = "Reykjavik, Iceland (Online)",
    publisher = {Link{\"o}ping University Electronic Press, Sweden},
    url = "https://aclanthology.org/2021.nodalida-main.6/",
    pages = "53--63"
}

@misc{grattafiori2024llama3herdmodels,
      title={The Llama 3 Herd of Models}, 
      author={Aaron Grattafiori and Abhimanyu Dubey and Abhinav Jauhri and Abhinav Pandey and Abhishek Kadian and Ahmad Al-Dahle and Aiesha Letman and Akhil Mathur and Alan Schelten and Alex Vaughan and Amy Yang and Angela Fan and Anirudh Goyal and Anthony Hartshorn and Aobo Yang and Archi Mitra and Archie Sravankumar and Artem Korenev and Arthur Hinsvark and Arun Rao and Aston Zhang and Aurelien Rodriguez and Austen Gregerson and Ava Spataru and Baptiste Roziere and Bethany Biron and Binh Tang and Bobbie Chern and Charlotte Caucheteux and Chaya Nayak and Chloe Bi and Chris Marra and Chris McConnell and Christian Keller and Christophe Touret and Chunyang Wu and Corinne Wong and Cristian Canton Ferrer and Cyrus Nikolaidis and Damien Allonsius and Daniel Song and Danielle Pintz and Danny Livshits and Danny Wyatt and David Esiobu and Dhruv Choudhary and Dhruv Mahajan and Diego Garcia-Olano and Diego Perino and Dieuwke Hupkes and Egor Lakomkin and Ehab AlBadawy and Elina Lobanova and Emily Dinan and Eric Michael Smith and Filip Radenovic and Francisco Guzmán and Frank Zhang and Gabriel Synnaeve and Gabrielle Lee and Georgia Lewis Anderson and Govind Thattai and Graeme Nail and Gregoire Mialon and Guan Pang and Guillem Cucurell and Hailey Nguyen and Hannah Korevaar and Hu Xu and Hugo Touvron and Iliyan Zarov and Imanol Arrieta Ibarra and Isabel Kloumann and Ishan Misra and Ivan Evtimov and Jack Zhang and Jade Copet and Jaewon Lee and Jan Geffert and Jana Vranes and Jason Park and Jay Mahadeokar and Jeet Shah and Jelmer van der Linde and Jennifer Billock and Jenny Hong and Jenya Lee and Jeremy Fu and Jianfeng Chi and Jianyu Huang and Jiawen Liu and Jie Wang and Jiecao Yu and Joanna Bitton and Joe Spisak and Jongsoo Park and Joseph Rocca and Joshua Johnstun and Joshua Saxe and Junteng Jia and Kalyan Vasuden Alwala and Karthik Prasad and Kartikeya Upasani and Kate Plawiak and Ke Li and Kenneth Heafield and Kevin Stone and Khalid El-Arini and Krithika Iyer and Kshitiz Malik and Kuenley Chiu and Kunal Bhalla and Kushal Lakhotia and Lauren Rantala-Yeary and Laurens van der Maaten and Lawrence Chen and Liang Tan and Liz Jenkins and Louis Martin and Lovish Madaan and Lubo Malo and Lukas Blecher and Lukas Landzaat and Luke de Oliveira and Madeline Muzzi and Mahesh Pasupuleti and Mannat Singh and Manohar Paluri and Marcin Kardas and Maria Tsimpoukelli and Mathew Oldham and Mathieu Rita and Maya Pavlova and Melanie Kambadur and Mike Lewis and Min Si and Mitesh Kumar Singh and Mona Hassan and Naman Goyal and Narjes Torabi and Nikolay Bashlykov and Nikolay Bogoychev and Niladri Chatterji and Ning Zhang and Olivier Duchenne and Onur Çelebi and Patrick Alrassy and Pengchuan Zhang and Pengwei Li and Petar Vasic and Peter Weng and Prajjwal Bhargava and Pratik Dubal and Praveen Krishnan and Punit Singh Koura and Puxin Xu and Qing He and Qingxiao Dong and Ragavan Srinivasan and Raj Ganapathy and Ramon Calderer and Ricardo Silveira Cabral and Robert Stojnic and Roberta Raileanu and Rohan Maheswari and Rohit Girdhar and Rohit Patel and Romain Sauvestre and Ronnie Polidoro and Roshan Sumbaly and Ross Taylor and Ruan Silva and Rui Hou and Rui Wang and Saghar Hosseini and Sahana Chennabasappa and Sanjay Singh and Sean Bell and Seohyun Sonia Kim and Sergey Edunov and Shaoliang Nie and Sharan Narang and Sharath Raparthy and Sheng Shen and Shengye Wan and Shruti Bhosale and Shun Zhang and Simon Vandenhende and Soumya Batra and Spencer Whitman and Sten Sootla and Stephane Collot and Suchin Gururangan and Sydney Borodinsky and Tamar Herman and Tara Fowler and Tarek Sheasha and Thomas Georgiou and Thomas Scialom and Tobias Speckbacher and Todor Mihaylov and Tong Xiao and Ujjwal Karn and Vedanuj Goswami and Vibhor Gupta and Vignesh Ramanathan and Viktor Kerkez and Vincent Gonguet and Virginie Do and Vish Vogeti and Vítor Albiero and Vladan Petrovic and Weiwei Chu and Wenhan Xiong and Wenyin Fu and Whitney Meers and Xavier Martinet and Xiaodong Wang and Xiaofang Wang and Xiaoqing Ellen Tan and Xide Xia and Xinfeng Xie and Xuchao Jia and Xuewei Wang and Yaelle Goldschlag and Yashesh Gaur and Yasmine Babaei and Yi Wen and Yiwen Song and Yuchen Zhang and Yue Li and Yuning Mao and Zacharie Delpierre Coudert and Zheng Yan and Zhengxing Chen and Zoe Papakipos and Aaditya Singh and Aayushi Srivastava and Abha Jain and Adam Kelsey and Adam Shajnfeld and Adithya Gangidi and Adolfo Victoria and Ahuva Goldstand and Ajay Menon and Ajay Sharma and Alex Boesenberg and Alexei Baevski and Allie Feinstein and Amanda Kallet and Amit Sangani and Amos Teo and Anam Yunus and Andrei Lupu and Andres Alvarado and Andrew Caples and Andrew Gu and Andrew Ho and Andrew Poulton and Andrew Ryan and Ankit Ramchandani and Annie Dong and Annie Franco and Anuj Goyal and Aparajita Saraf and Arkabandhu Chowdhury and Ashley Gabriel and Ashwin Bharambe and Assaf Eisenman and Azadeh Yazdan and Beau James and Ben Maurer and Benjamin Leonhardi and Bernie Huang and Beth Loyd and Beto De Paola and Bhargavi Paranjape and Bing Liu and Bo Wu and Boyu Ni and Braden Hancock and Bram Wasti and Brandon Spence and Brani Stojkovic and Brian Gamido and Britt Montalvo and Carl Parker and Carly Burton and Catalina Mejia and Ce Liu and Changhan Wang and Changkyu Kim and Chao Zhou and Chester Hu and Ching-Hsiang Chu and Chris Cai and Chris Tindal and Christoph Feichtenhofer and Cynthia Gao and Damon Civin and Dana Beaty and Daniel Kreymer and Daniel Li and David Adkins and David Xu and Davide Testuggine and Delia David and Devi Parikh and Diana Liskovich and Didem Foss and Dingkang Wang and Duc Le and Dustin Holland and Edward Dowling and Eissa Jamil and Elaine Montgomery and Eleonora Presani and Emily Hahn and Emily Wood and Eric-Tuan Le and Erik Brinkman and Esteban Arcaute and Evan Dunbar and Evan Smothers and Fei Sun and Felix Kreuk and Feng Tian and Filippos Kokkinos and Firat Ozgenel and Francesco Caggioni and Frank Kanayet and Frank Seide and Gabriela Medina Florez and Gabriella Schwarz and Gada Badeer and Georgia Swee and Gil Halpern and Grant Herman and Grigory Sizov and Guangyi and Zhang and Guna Lakshminarayanan and Hakan Inan and Hamid Shojanazeri and Han Zou and Hannah Wang and Hanwen Zha and Haroun Habeeb and Harrison Rudolph and Helen Suk and Henry Aspegren and Hunter Goldman and Hongyuan Zhan and Ibrahim Damlaj and Igor Molybog and Igor Tufanov and Ilias Leontiadis and Irina-Elena Veliche and Itai Gat and Jake Weissman and James Geboski and James Kohli and Janice Lam and Japhet Asher and Jean-Baptiste Gaya and Jeff Marcus and Jeff Tang and Jennifer Chan and Jenny Zhen and Jeremy Reizenstein and Jeremy Teboul and Jessica Zhong and Jian Jin and Jingyi Yang and Joe Cummings and Jon Carvill and Jon Shepard and Jonathan McPhie and Jonathan Torres and Josh Ginsburg and Junjie Wang and Kai Wu and Kam Hou U and Karan Saxena and Kartikay Khandelwal and Katayoun Zand and Kathy Matosich and Kaushik Veeraraghavan and Kelly Michelena and Keqian Li and Kiran Jagadeesh and Kun Huang and Kunal Chawla and Kyle Huang and Lailin Chen and Lakshya Garg and Lavender A and Leandro Silva and Lee Bell and Lei Zhang and Liangpeng Guo and Licheng Yu and Liron Moshkovich and Luca Wehrstedt and Madian Khabsa and Manav Avalani and Manish Bhatt and Martynas Mankus and Matan Hasson and Matthew Lennie and Matthias Reso and Maxim Groshev and Maxim Naumov and Maya Lathi and Meghan Keneally and Miao Liu and Michael L. Seltzer and Michal Valko and Michelle Restrepo and Mihir Patel and Mik Vyatskov and Mikayel Samvelyan and Mike Clark and Mike Macey and Mike Wang and Miquel Jubert Hermoso and Mo Metanat and Mohammad Rastegari and Munish Bansal and Nandhini Santhanam and Natascha Parks and Natasha White and Navyata Bawa and Nayan Singhal and Nick Egebo and Nicolas Usunier and Nikhil Mehta and Nikolay Pavlovich Laptev and Ning Dong and Norman Cheng and Oleg Chernoguz and Olivia Hart and Omkar Salpekar and Ozlem Kalinli and Parkin Kent and Parth Parekh and Paul Saab and Pavan Balaji and Pedro Rittner and Philip Bontrager and Pierre Roux and Piotr Dollar and Polina Zvyagina and Prashant Ratanchandani and Pritish Yuvraj and Qian Liang and Rachad Alao and Rachel Rodriguez and Rafi Ayub and Raghotham Murthy and Raghu Nayani and Rahul Mitra and Rangaprabhu Parthasarathy and Raymond Li and Rebekkah Hogan and Robin Battey and Rocky Wang and Russ Howes and Ruty Rinott and Sachin Mehta and Sachin Siby and Sai Jayesh Bondu and Samyak Datta and Sara Chugh and Sara Hunt and Sargun Dhillon and Sasha Sidorov and Satadru Pan and Saurabh Mahajan and Saurabh Verma and Seiji Yamamoto and Sharadh Ramaswamy and Shaun Lindsay and Shaun Lindsay and Sheng Feng and Shenghao Lin and Shengxin Cindy Zha and Shishir Patil and Shiva Shankar and Shuqiang Zhang and Shuqiang Zhang and Sinong Wang and Sneha Agarwal and Soji Sajuyigbe and Soumith Chintala and Stephanie Max and Stephen Chen and Steve Kehoe and Steve Satterfield and Sudarshan Govindaprasad and Sumit Gupta and Summer Deng and Sungmin Cho and Sunny Virk and Suraj Subramanian and Sy Choudhury and Sydney Goldman and Tal Remez and Tamar Glaser and Tamara Best and Thilo Koehler and Thomas Robinson and Tianhe Li and Tianjun Zhang and Tim Matthews and Timothy Chou and Tzook Shaked and Varun Vontimitta and Victoria Ajayi and Victoria Montanez and Vijai Mohan and Vinay Satish Kumar and Vishal Mangla and Vlad Ionescu and Vlad Poenaru and Vlad Tiberiu Mihailescu and Vladimir Ivanov and Wei Li and Wenchen Wang and Wenwen Jiang and Wes Bouaziz and Will Constable and Xiaocheng Tang and Xiaojian Wu and Xiaolan Wang and Xilun Wu and Xinbo Gao and Yaniv Kleinman and Yanjun Chen and Ye Hu and Ye Jia and Ye Qi and Yenda Li and Yilin Zhang and Ying Zhang and Yossi Adi and Youngjin Nam and Yu and Wang and Yu Zhao and Yuchen Hao and Yundi Qian and Yunlu Li and Yuzi He and Zach Rait and Zachary DeVito and Zef Rosnbrick and Zhaoduo Wen and Zhenyu Yang and Zhiwei Zhao and Zhiyu Ma},
      year={2024},
      eprint={2407.21783},
      archivePrefix={arXiv},
      primaryClass={cs.AI},
      url={https://arxiv.org/abs/2407.21783}, 
}

@inproceedings{ermis-etal-2024-one,
    title = "From One to Many: Expanding the Scope of Toxicity Mitigation in Language Models",
    author = "Ermis, Beyza  and
      Pozzobon, Luiza  and
      Hooker, Sara  and
      Lewis, Patrick",
    editor = "Ku, Lun-Wei  and
      Martins, Andre  and
      Srikumar, Vivek",
    booktitle = "Findings of the Association for Computational Linguistics: ACL 2024",
    month = aug,
    year = "2024",
    address = "Bangkok, Thailand",
    publisher = "Association for Computational Linguistics",
    url = "https://aclanthology.org/2024.findings-acl.893/",
    doi = "10.18653/v1/2024.findings-acl.893",
    pages = "15041--15058"
}

@inproceedings{marchisio-etal-2024-understanding,
    title = "Understanding and Mitigating Language Confusion in {LLM}s",
    author = "Marchisio, Kelly  and
      Ko, Wei-Yin  and
      Berard, Alexandre  and
      Dehaze, Th{\'e}o  and
      Ruder, Sebastian",
    editor = "Al-Onaizan, Yaser  and
      Bansal, Mohit  and
      Chen, Yun-Nung",
    booktitle = "Proceedings of the 2024 Conference on Empirical Methods in Natural Language Processing",
    month = nov,
    year = "2024",
    address = "Miami, Florida, USA",
    publisher = "Association for Computational Linguistics",
    url = "https://aclanthology.org/2024.emnlp-main.380/",
    doi = "10.18653/v1/2024.emnlp-main.380",
    pages = "6653--6677"
}

@inproceedings{
kreutzer2025dj,
title={D\'ej\`a Vu: Multilingual {LLM} Evaluation through the Lens of Machine Translation Evaluation},
author={Julia Kreutzer and Eleftheria Briakou and Sweta Agrawal and Marzieh Fadaee and Tom Kocmi},
booktitle={Second Conference on Language Modeling},
year={2025},
url={https://openreview.net/forum?id=yxzVanFoij}
}

@misc{khairi2025makingtakingbestn,
      title={Making, not Taking, the Best of N}, 
      author={Ammar Khairi and Daniel D'souza and Marzieh Fadaee and Julia Kreutzer},
      year={2025},
      eprint={2510.00931},
      archivePrefix={arXiv},
      primaryClass={cs.CL},
      url={https://arxiv.org/abs/2510.00931}, 
}

@inproceedings{thompson-etal-2024-shocking,
    title = "A Shocking Amount of the Web is Machine Translated: Insights from Multi-Way Parallelism",
    author = "Thompson, Brian  and
      Dhaliwal, Mehak  and
      Frisch, Peter  and
      Domhan, Tobias  and
      Federico, Marcello",
    editor = "Ku, Lun-Wei  and
      Martins, Andre  and
      Srikumar, Vivek",
    booktitle = "Findings of the Association for Computational Linguistics: ACL 2024",
    month = aug,
    year = "2024",
    address = "Bangkok, Thailand",
    publisher = "Association for Computational Linguistics",
    url = "https://aclanthology.org/2024.findings-acl.103/",
    doi = "10.18653/v1/2024.findings-acl.103",
    pages = "1763--1775"
}

@misc{schreiter2025promptengineeringpromptvocabulary,
      title={Prompt Engineering: How Prompt Vocabulary affects Domain Knowledge}, 
      author={Dimitri Schreiter},
      year={2025},
      eprint={2505.17037},
      archivePrefix={arXiv},
      primaryClass={cs.CL},
      url={https://arxiv.org/abs/2505.17037}, 
}

@inproceedings{
ajayi2026what,
title={What Do Prompts Reveal About Model Capabilities in Low-Resource Languages?},
author={Oluwaseun A. Ajayi},
booktitle={7th Workshop on African Natural Language Processing},
year={2026},
url={https://openreview.net/forum?id=7JZmTp85Yf}
}

@misc{opsahlong2024miprov2,
  title={Optimizing Instructions and Demonstrations for Multi-Stage Language Model Programs},
  author={Krista Opsahl-Ong and Michael J. Ryan and Josh Purtell and David Broman and Christopher Potts and Matei Zaharia and Omar Khattab},
  year={2024},
  eprint={2406.11695},
  archivePrefix={arXiv},
  primaryClass={cs.CL},
  url={https://arxiv.org/abs/2406.11695}
}

@misc{chen2024mapo,
  title={MAPO: Boosting Large Language Model Performance with Model-Adaptive Prompt Optimization},
  author={Yuyan Chen and Zhihao Wen and Ge Fan and Zhengyu Chen and Wei Wu and Dayiheng Liu and Zhixu Li and Bang Liu and Yanghua Xiao},
  year={2024},
  eprint={2407.04118},
  archivePrefix={arXiv},
  primaryClass={cs.CL},
  url={https://arxiv.org/abs/2407.04118}
}

@misc{agrawal2025gepa,
  title={GEPA: Reflective Prompt Evolution Can Outperform Reinforcement Learning},
  author={Lakshya A. Agrawal and Shangyin Tan and Dilara Soylu and Noah Ziems and Rishi Khare and Krista Opsahl-Ong and Arnav Singhvi and Herumb Shandilya and Michael J. Ryan and Meng Jiang and Christopher Potts and Koushik Sen and Alexandros G. Dimakis and Ion Stoica and Dan Klein and Matei Zaharia and Omar Khattab},
  year={2025},
  eprint={2507.19457},
  archivePrefix={arXiv},
  primaryClass={cs.CL},
  url={https://arxiv.org/abs/2507.19457}
}

@misc{huang2025benchmaxcomprehensivemultilingualevaluation,
      title={BenchMAX: A Comprehensive Multilingual Evaluation Suite for Large Language Models}, 
      author={Xu Huang and Wenhao Zhu and Hanxu Hu and Conghui He and Lei Li and Shujian Huang and Fei Yuan},
      year={2025},
      eprint={2502.07346},
      archivePrefix={arXiv},
      primaryClass={cs.CL},
      url={https://arxiv.org/abs/2502.07346}, 
}

@inproceedings{baucells-etal-2025-iberobench,
    title = "{I}bero{B}ench: A Benchmark for {LLM} Evaluation in {I}berian Languages",
    author = "Baucells, Irene  and
      Aula-Blasco, Javier  and
      de-Dios-Flores, Iria  and
      Paniagua Su{\'a}rez, Silvia  and
      Perez, Naiara  and
      Salles, Anna  and
      Sotelo Docio, Susana  and
      Falc{\~a}o, J{\'u}lia  and
      Saiz, Jose Javier  and
      Sepulveda Torres, Robiert  and
      Barnes, Jeremy  and
      Gamallo, Pablo  and
      Gonzalez-Agirre, Aitor  and
      Rigau, German  and
      Villegas, Marta",
    editor = "Rambow, Owen  and
      Wanner, Leo  and
      Apidianaki, Marianna  and
      Al-Khalifa, Hend  and
      Eugenio, Barbara Di  and
      Schockaert, Steven",
    booktitle = "Proceedings of the 31st International Conference on Computational Linguistics",
    month = jan,
    year = "2025",
    address = "Abu Dhabi, UAE",
    publisher = "Association for Computational Linguistics",
    url = "https://aclanthology.org/2025.coling-main.699/",
    pages = "10491--10519"
}

@inproceedings{ranathunga-de-silva-2022-languages,
    title = "Some Languages are More Equal than Others: Probing Deeper into the Linguistic Disparity in the {NLP} World",
    author = "Ranathunga, Surangika  and
      de Silva, Nisansa",
    editor = "He, Yulan  and
      Ji, Heng  and
      Li, Sujian  and
      Liu, Yang  and
      Chang, Chua-Hui",
    booktitle = "Proceedings of the 2nd Conference of the Asia-Pacific Chapter of the Association for Computational Linguistics and the 12th International Joint Conference on Natural Language Processing (Volume 1: Long Papers)",
    month = nov,
    year = "2022",
    address = "Online only",
    publisher = "Association for Computational Linguistics",
    url = "https://aclanthology.org/2022.aacl-main.62/",
    doi = "10.18653/v1/2022.aacl-main.62",
    pages = "823--848"
}

@inproceedings{bird-2022-local,
    title = "Local Languages, Third Spaces, and other High-Resource Scenarios",
    author = "Bird, Steven",
    editor = "Muresan, Smaranda  and
      Nakov, Preslav  and
      Villavicencio, Aline",
    booktitle = "Proceedings of the 60th Annual Meeting of the Association for Computational Linguistics (Volume 1: Long Papers)",
    month = may,
    year = "2022",
    address = "Dublin, Ireland",
    publisher = "Association for Computational Linguistics",
    url = "https://aclanthology.org/2022.acl-long.539/",
    doi = "10.18653/v1/2022.acl-long.539",
    pages = "7817--7829"
}

@inproceedings{enomoto-etal-2025-fair,
    title = "A Fair Comparison without Translationese: {E}nglish vs. Target-language Instructions for Multilingual {LLM}s",
    author = "Enomoto, Taisei  and
      Kim, Hwichan  and
      Chen, Zhousi  and
      Komachi, Mamoru",
    editor = "Chiruzzo, Luis  and
      Ritter, Alan  and
      Wang, Lu",
    booktitle = "Proceedings of the 2025 Conference of the Nations of the Americas Chapter of the Association for Computational Linguistics: Human Language Technologies (Volume 2: Short Papers)",
    month = apr,
    year = "2025",
    address = "Albuquerque, New Mexico",
    publisher = "Association for Computational Linguistics",
    url = "https://aclanthology.org/2025.naacl-short.55/",
    doi = "10.18653/v1/2025.naacl-short.55",
    pages = "649--670",
    ISBN = "979-8-89176-190-2"
}

@inproceedings{eetemadi-toutanova-2014-asymmetric,
    title = "Asymmetric Features Of Human Generated Translation",
    author = "Eetemadi, Sauleh  and
      Toutanova, Kristina",
    editor = "Moschitti, Alessandro  and
      Pang, Bo  and
      Daelemans, Walter",
    booktitle = "Proceedings of the 2014 Conference on Empirical Methods in Natural Language Processing ({EMNLP})",
    month = oct,
    year = "2014",
    address = "Doha, Qatar",
    publisher = "Association for Computational Linguistics",
    url = "https://aclanthology.org/D14-1018/",
    doi = "10.3115/v1/D14-1018",
    pages = "159--164"
}

@misc{hernándezcano2025apertusdemocratizingopencompliant,
      title={Apertus: Democratizing Open and Compliant LLMs for Global Language Environments}, 
      author={Alejandro Hernández-Cano and Alexander Hägele and Allen Hao Huang and Angelika Romanou and Antoni-Joan Solergibert and Barna Pasztor and Bettina Messmer and Dhia Garbaya and Eduard Frank Ďurech and Ido Hakimi and Juan García Giraldo and Mete Ismayilzada and Negar Foroutan and Skander Moalla and Tiancheng Chen and Vinko Sabolčec and Yixuan Xu and Michael Aerni and Badr AlKhamissi and Ines Altemir Marinas and Mohammad Hossein Amani and Matin Ansaripour and Ilia Badanin and Harold Benoit and Emanuela Boros and Nicholas Browning and Fabian Bösch and Maximilian Böther and Niklas Canova and Camille Challier and Clement Charmillot and Jonathan Coles and Jan Deriu and Arnout Devos and Lukas Drescher and Daniil Dzenhaliou and Maud Ehrmann and Dongyang Fan and Simin Fan and Silin Gao and Miguel Gila and María Grandury and Diba Hashemi and Alexander Hoyle and Jiaming Jiang and Mark Klein and Andrei Kucharavy and Anastasiia Kucherenko and Frederike Lübeck and Roman Machacek and Theofilos Manitaras and Andreas Marfurt and Kyle Matoba and Simon Matrenok and Henrique Mendoncça and Fawzi Roberto Mohamed and Syrielle Montariol and Luca Mouchel and Sven Najem-Meyer and Jingwei Ni and Gennaro Oliva and Matteo Pagliardini and Elia Palme and Andrei Panferov and Léo Paoletti and Marco Passerini and Ivan Pavlov and Auguste Poiroux and Kaustubh Ponkshe and Nathan Ranchin and Javi Rando and Mathieu Sauser and Jakhongir Saydaliev and Muhammad Ali Sayfiddinov and Marian Schneider and Stefano Schuppli and Marco Scialanga and Andrei Semenov and Kumar Shridhar and Raghav Singhal and Anna Sotnikova and Alexander Sternfeld and Ayush Kumar Tarun and Paul Teiletche and Jannis Vamvas and Xiaozhe Yao and Hao Zhao Alexander Ilic and Ana Klimovic and Andreas Krause and Caglar Gulcehre and David Rosenthal and Elliott Ash and Florian Tramèr and Joost VandeVondele and Livio Veraldi and Martin Rajman and Thomas Schulthess and Torsten Hoefler and Antoine Bosselut and Martin Jaggi and Imanol Schlag},
      year={2025},
      eprint={2509.14233},
      archivePrefix={arXiv},
      primaryClass={cs.CL},
      url={https://arxiv.org/abs/2509.14233}, 
}

@misc{martins2025eurollm9btechnicalreport,
      title={EuroLLM-9B: Technical Report}, 
      author={Pedro Henrique Martins and João Alves and Patrick Fernandes and Nuno M. Guerreiro and Ricardo Rei and Amin Farajian and Mateusz Klimaszewski and Duarte M. Alves and José Pombal and Nicolas Boizard and Manuel Faysse and Pierre Colombo and François Yvon and Barry Haddow and José G. C. de Souza and Alexandra Birch and André F. T. Martins},
      year={2025},
      eprint={2506.04079},
      archivePrefix={arXiv},
      primaryClass={cs.CL},
      url={https://arxiv.org/abs/2506.04079}, 
}

@inproceedings{li-etal-2025-lost,
    title = "Lost in Literalism: How Supervised Training Shapes Translationese in {LLM}s",
    author = "Li, Yafu  and
      Zhang, Ronghao  and
      Wang, Zhilin  and
      Zhang, Huajian  and
      Cui, Leyang  and
      Yin, Yongjing  and
      Xiao, Tong  and
      Zhang, Yue",
    editor = "Che, Wanxiang  and
      Nabende, Joyce  and
      Shutova, Ekaterina  and
      Pilehvar, Mohammad Taher",
    booktitle = "Proceedings of the 63rd Annual Meeting of the Association for Computational Linguistics (Volume 1: Long Papers)",
    month = jul,
    year = "2025",
    address = "Vienna, Austria",
    publisher = "Association for Computational Linguistics",
    url = "https://aclanthology.org/2025.acl-long.630/",
    doi = "10.18653/v1/2025.acl-long.630",
    pages = "12875--12894",
    ISBN = "979-8-89176-251-0"
}

@article{nllb2022,
  author    = {NLLB Team and Marta R. Costa-jussà and James Cross and Onur Çelebi and Maha Elbayad and Kenneth Heafield and Kevin Heffernan and Elahe Kalbassi and Janice Lam and Daniel Licht and Jean Maillard and Anna Sun and Skyler Wang and Guillaume Wenzek and Al Youngblood and Bapi Akula and Loic Barrault and Gabriel Mejia Gonzalez and Prangthip Hansanti and John Hoffman and Semarley Jarrett and Kaushik Ram Sadagopan and Dirk Rowe and Shannon Spruit and Chau Tran and Pierre Andrews and Necip Fazil Ayan and Shruti Bhosale and Sergey Edunov and Angela Fan and Cynthia Gao and Vedanuj Goswami and Francisco Guzmán and Philipp Koehn and Alexandre Mourachko and Christophe Ropers and Safiyyah Saleem and Holger Schwenk and Jeff Wang},
  title     = {No Language Left Behind: Scaling Human-Centered Machine Translation},
  year      = {2022}
}

@inproceedings{
liu2024best,
title={Best Practices and Lessons Learned on Synthetic Data},
author={Ruibo Liu and Jerry Wei and Fangyu Liu and Chenglei Si and Yanzhe Zhang and Jinmeng Rao and Steven Zheng and Daiyi Peng and Diyi Yang and Denny Zhou and Andrew M. Dai},
booktitle={First Conference on Language Modeling},
year={2024},
url={https://openreview.net/forum?id=OJaWBhh61C}
}

@inproceedings{long-etal-2024-llms,
    title = "On {LLM}s-Driven Synthetic Data Generation, Curation, and Evaluation: A Survey",
    author = "Long, Lin  and
      Wang, Rui  and
      Xiao, Ruixuan  and
      Zhao, Junbo  and
      Ding, Xiao  and
      Chen, Gang  and
      Wang, Haobo",
    editor = "Ku, Lun-Wei  and
      Martins, Andre  and
      Srikumar, Vivek",
    booktitle = "Findings of the Association for Computational Linguistics: ACL 2024",
    month = aug,
    year = "2024",
    address = "Bangkok, Thailand",
    publisher = "Association for Computational Linguistics",
    url = "https://aclanthology.org/2024.findings-acl.658/",
    doi = "10.18653/v1/2024.findings-acl.658",
    pages = "11065--11082"
}

@misc{khairi2025lifegivessamplesbenefits,
      title={When Life Gives You Samples: The Benefits of Scaling up Inference Compute for Multilingual LLMs}, 
      author={Ammar Khairi and Daniel D'souza and Ye Shen and Julia Kreutzer and Sara Hooker},
      year={2025},
      eprint={2506.20544},
      archivePrefix={arXiv},
      primaryClass={cs.CL},
      url={https://arxiv.org/abs/2506.20544}, 
}

@inproceedings{shimabucoro-etal-2024-llm,
    title = "{LLM} See, {LLM} Do: Leveraging Active Inheritance to Target Non-Differentiable Objectives",
    author = "Shimabucoro, Lu{\'i}sa  and
      Ruder, Sebastian  and
      Kreutzer, Julia  and
      Fadaee, Marzieh  and
      Hooker, Sara",
    editor = "Al-Onaizan, Yaser  and
      Bansal, Mohit  and
      Chen, Yun-Nung",
    booktitle = "Proceedings of the 2024 Conference on Empirical Methods in Natural Language Processing",
    month = nov,
    year = "2024",
    address = "Miami, Florida, USA",
    publisher = "Association for Computational Linguistics",
    url = "https://aclanthology.org/2024.emnlp-main.521/",
    doi = "10.18653/v1/2024.emnlp-main.521",
    pages = "9243--9267"
}

@inproceedings{ustun-etal-2024-aya,
    title = "Aya Model: An Instruction Finetuned Open-Access Multilingual Language Model",
    author = {{\"U}st{\"u}n, Ahmet  and
      Aryabumi, Viraat  and
      Yong, Zheng  and
      Ko, Wei-Yin  and
      D{'}souza, Daniel  and
      Onilude, Gbemileke  and
      Bhandari, Neel  and
      Singh, Shivalika  and
      Ooi, Hui-Lee  and
      Kayid, Amr  and
      Vargus, Freddie  and
      Blunsom, Phil  and
      Longpre, Shayne  and
      Muennighoff, Niklas  and
      Fadaee, Marzieh  and
      Kreutzer, Julia  and
      Hooker, Sara},
    editor = "Ku, Lun-Wei  and
      Martins, Andre  and
      Srikumar, Vivek",
    booktitle = "Proceedings of the 62nd Annual Meeting of the Association for Computational Linguistics (Volume 1: Long Papers)",
    month = aug,
    year = "2024",
    address = "Bangkok, Thailand",
    publisher = "Association for Computational Linguistics",
    url = "https://aclanthology.org/2024.acl-long.845/",
    doi = "10.18653/v1/2024.acl-long.845",
    pages = "15894--15939"
}

@misc{dang2024ayaexpansecombiningresearch,
      title={Aya Expanse: Combining Research Breakthroughs for a New Multilingual Frontier}, 
      author={John Dang and Shivalika Singh and Daniel D'souza and Arash Ahmadian and Alejandro Salamanca and Madeline Smith and Aidan Peppin and Sungjin Hong and Manoj Govindassamy and Terrence Zhao and Sandra Kublik and Meor Amer and Viraat Aryabumi and Jon Ander Campos and Yi-Chern Tan and Tom Kocmi and Florian Strub and Nathan Grinsztajn and Yannis Flet-Berliac and Acyr Locatelli and Hangyu Lin and Dwarak Talupuru and Bharat Venkitesh and David Cairuz and Bowen Yang and Tim Chung and Wei-Yin Ko and Sylvie Shang Shi and Amir Shukayev and Sammie Bae and Aleksandra Piktus and Roman Castagné and Felipe Cruz-Salinas and Eddie Kim and Lucas Crawhall-Stein and Adrien Morisot and Sudip Roy and Phil Blunsom and Ivan Zhang and Aidan Gomez and Nick Frosst and Marzieh Fadaee and Beyza Ermis and Ahmet Üstün and Sara Hooker},
      year={2024},
      eprint={2412.04261},
      archivePrefix={arXiv},
      primaryClass={cs.CL},
      url={https://arxiv.org/abs/2412.04261}, 
}

@inproceedings{chen-etal-2024-good-data,
    title = "Is It Good Data for Multilingual Instruction Tuning or Just Bad Multilingual Evaluation for Large Language Models?",
    author = "Chen, Pinzhen  and
      Yu, Simon  and
      Guo, Zhicheng  and
      Haddow, Barry",
    editor = "Al-Onaizan, Yaser  and
      Bansal, Mohit  and
      Chen, Yun-Nung",
    booktitle = "Proceedings of the 2024 Conference on Empirical Methods in Natural Language Processing",
    month = nov,
    year = "2024",
    address = "Miami, Florida, USA",
    publisher = "Association for Computational Linguistics",
    url = "https://aclanthology.org/2024.emnlp-main.542/",
    doi = "10.18653/v1/2024.emnlp-main.542",
    pages = "9706--9726"
}

@article{10.1162/COLI_a_00111,
    author = {Lembersky, Gennadi and Ordan, Noam and Wintner, Shuly},
    title = {Language Models for Machine Translation: Original vs. Translated Texts },
    journal = {Computational Linguistics},
    volume = {38},
    number = {4},
    pages = {799-825},
    year = {2012},
    month = {12},
    issn = {0891-2017},
    doi = {10.1162/COLI_a_00111},
    url = {https://doi.org/10.1162/COLI_a_00111},
    eprint = {https://direct.mit.edu/coli/article-pdf/38/4/799/1799785/coli_a_00111.pdf},
}

@misc{romanou2024includeevaluatingmultilinguallanguage,
      title={INCLUDE: Evaluating Multilingual Language Understanding with Regional Knowledge}, 
      author={Angelika Romanou and Negar Foroutan and Anna Sotnikova and Zeming Chen and Sree Harsha Nelaturu and Shivalika Singh and Rishabh Maheshwary and Micol Altomare and Mohamed A. Haggag and Snegha A and Alfonso Amayuelas and Azril Hafizi Amirudin and Viraat Aryabumi and Danylo Boiko and Michael Chang and Jenny Chim and Gal Cohen and Aditya Kumar Dalmia and Abraham Diress and Sharad Duwal and Daniil Dzenhaliou and Daniel Fernando Erazo Florez and Fabian Farestam and Joseph Marvin Imperial and Shayekh Bin Islam and Perttu Isotalo and Maral Jabbarishiviari and Börje F. Karlsson and Eldar Khalilov and Christopher Klamm and Fajri Koto and Dominik Krzemiński and Gabriel Adriano de Melo and Syrielle Montariol and Yiyang Nan and Joel Niklaus and Jekaterina Novikova and Johan Samir Obando Ceron and Debjit Paul and Esther Ploeger and Jebish Purbey and Swati Rajwal and Selvan Sunitha Ravi and Sara Rydell and Roshan Santhosh and Drishti Sharma and Marjana Prifti Skenduli and Arshia Soltani Moakhar and Bardia Soltani Moakhar and Ran Tamir and Ayush Kumar Tarun and Azmine Toushik Wasi and Thenuka Ovin Weerasinghe and Serhan Yilmaz and Mike Zhang and Imanol Schlag and Marzieh Fadaee and Sara Hooker and Antoine Bosselut},
      year={2024},
      eprint={2411.19799},
      archivePrefix={arXiv},
      primaryClass={cs.CL},
      url={https://arxiv.org/abs/2411.19799}, 
}

@inproceedings{singh-etal-2025-global,
    title = "Global {MMLU}: Understanding and Addressing Cultural and Linguistic Biases in Multilingual Evaluation",
    author = "Singh, Shivalika  and
      Romanou, Angelika  and
      Fourrier, Cl{\'e}mentine  and
      Adelani, David Ifeoluwa  and
      Ngui, Jian Gang  and
      Vila-Suero, Daniel  and
      Limkonchotiwat, Peerat  and
      Marchisio, Kelly  and
      Leong, Wei Qi  and
      Susanto, Yosephine  and
      Ng, Raymond  and
      Longpre, Shayne  and
      Ruder, Sebastian  and
      Ko, Wei-Yin  and
      Bosselut, Antoine  and
      Oh, Alice  and
      Martins, Andre  and
      Choshen, Leshem  and
      Ippolito, Daphne  and
      Ferrante, Enzo  and
      Fadaee, Marzieh  and
      Ermis, Beyza  and
      Hooker, Sara",
    editor = "Che, Wanxiang  and
      Nabende, Joyce  and
      Shutova, Ekaterina  and
      Pilehvar, Mohammad Taher",
    booktitle = "Proceedings of the 63rd Annual Meeting of the Association for Computational Linguistics (Volume 1: Long Papers)",
    month = jul,
    year = "2025",
    address = "Vienna, Austria",
    publisher = "Association for Computational Linguistics",
    url = "https://aclanthology.org/2025.acl-long.919/",
    doi = "10.18653/v1/2025.acl-long.919",
    pages = "18761--18799",
    ISBN = "979-8-89176-251-0"
}

@inproceedings{deutsch-etal-2025-wmt24,
    title = "{WMT}24++: Expanding the Language Coverage of {WMT}24 to 55 Languages {\&} Dialects",
    author = "Deutsch, Daniel  and
      Briakou, Eleftheria  and
      Caswell, Isaac Rayburn  and
      Finkelstein, Mara  and
      Galor, Rebecca  and
      Juraska, Juraj  and
      Kovacs, Geza  and
      Lui, Alison  and
      Rei, Ricardo  and
      Riesa, Jason  and
      Rijhwani, Shruti  and
      Riley, Parker  and
      Salesky, Elizabeth  and
      Trabelsi, Firas  and
      Winkler, Stephanie  and
      Zhang, Biao  and
      Freitag, Markus",
    editor = "Che, Wanxiang  and
      Nabende, Joyce  and
      Shutova, Ekaterina  and
      Pilehvar, Mohammad Taher",
    booktitle = "Findings of the Association for Computational Linguistics: ACL 2025",
    month = jul,
    year = "2025",
    address = "Vienna, Austria",
    publisher = "Association for Computational Linguistics",
    url = "https://aclanthology.org/2025.findings-acl.634/",
    doi = "10.18653/v1/2025.findings-acl.634",
    pages = "12257--12284",
    ISBN = "979-8-89176-256-5"
}

@article{guerreiro-etal-2024-xcomet,
    title = "xcomet: Transparent Machine Translation Evaluation through Fine-grained Error Detection",
    author = "Guerreiro, Nuno M.  and
      Rei, Ricardo  and
      Stigt, Daan van  and
      Coheur, Luisa  and
      Colombo, Pierre  and
      Martins, Andr{\'e} F. T.",
    journal = "Transactions of the Association for Computational Linguistics",
    volume = "12",
    year = "2024",
    address = "Cambridge, MA",
    publisher = "MIT Press",
    url = "https://aclanthology.org/2024.tacl-1.54/",
    doi = "10.1162/tacl_a_00683",
    pages = "979--995"
}

@inproceedings{odumakinde-etal-2025-multilingual,
    title = "Multilingual Arbitration: Optimizing Data Pools to Accelerate Multilingual Progress",
    author = "Odumakinde, Ayomide  and
      D{'}souza, Daniel  and
      Verga, Pat  and
      Ermis, Beyza  and
      Hooker, Sara",
    editor = "Che, Wanxiang  and
      Nabende, Joyce  and
      Shutova, Ekaterina  and
      Pilehvar, Mohammad Taher",
    booktitle = "Proceedings of the 63rd Annual Meeting of the Association for Computational Linguistics (Volume 1: Long Papers)",
    month = jul,
    year = "2025",
    address = "Vienna, Austria",
    publisher = "Association for Computational Linguistics",
    url = "https://aclanthology.org/2025.acl-long.939/",
    doi = "10.18653/v1/2025.acl-long.939",
    pages = "19142--19164",
    ISBN = "979-8-89176-251-0"
}

@misc{gemmateam2025gemma3technicalreport,
      title={Gemma 3 Technical Report}, 
      author={Gemma Team and Aishwarya Kamath and Johan Ferret and Shreya Pathak and Nino Vieillard and Ramona Merhej and Sarah Perrin and Tatiana Matejovicova and Alexandre Ramé and Morgane Rivière and Louis Rouillard and Thomas Mesnard and Geoffrey Cideron and Jean-bastien Grill and Sabela Ramos and Edouard Yvinec and Michelle Casbon and Etienne Pot and Ivo Penchev and Gaël Liu and Francesco Visin and Kathleen Kenealy and Lucas Beyer and Xiaohai Zhai and Anton Tsitsulin and Robert Busa-Fekete and Alex Feng and Noveen Sachdeva and Benjamin Coleman and Yi Gao and Basil Mustafa and Iain Barr and Emilio Parisotto and David Tian and Matan Eyal and Colin Cherry and Jan-Thorsten Peter and Danila Sinopalnikov and Surya Bhupatiraju and Rishabh Agarwal and Mehran Kazemi and Dan Malkin and Ravin Kumar and David Vilar and Idan Brusilovsky and Jiaming Luo and Andreas Steiner and Abe Friesen and Abhanshu Sharma and Abheesht Sharma and Adi Mayrav Gilady and Adrian Goedeckemeyer and Alaa Saade and Alex Feng and Alexander Kolesnikov and Alexei Bendebury and Alvin Abdagic and Amit Vadi and András György and André Susano Pinto and Anil Das and Ankur Bapna and Antoine Miech and Antoine Yang and Antonia Paterson and Ashish Shenoy and Ayan Chakrabarti and Bilal Piot and Bo Wu and Bobak Shahriari and Bryce Petrini and Charlie Chen and Charline Le Lan and Christopher A. Choquette-Choo and CJ Carey and Cormac Brick and Daniel Deutsch and Danielle Eisenbud and Dee Cattle and Derek Cheng and Dimitris Paparas and Divyashree Shivakumar Sreepathihalli and Doug Reid and Dustin Tran and Dustin Zelle and Eric Noland and Erwin Huizenga and Eugene Kharitonov and Frederick Liu and Gagik Amirkhanyan and Glenn Cameron and Hadi Hashemi and Hanna Klimczak-Plucińska and Harman Singh and Harsh Mehta and Harshal Tushar Lehri and Hussein Hazimeh and Ian Ballantyne and Idan Szpektor and Ivan Nardini and Jean Pouget-Abadie and Jetha Chan and Joe Stanton and John Wieting and Jonathan Lai and Jordi Orbay and Joseph Fernandez and Josh Newlan and Ju-yeong Ji and Jyotinder Singh and Kat Black and Kathy Yu and Kevin Hui and Kiran Vodrahalli and Klaus Greff and Linhai Qiu and Marcella Valentine and Marina Coelho and Marvin Ritter and Matt Hoffman and Matthew Watson and Mayank Chaturvedi and Michael Moynihan and Min Ma and Nabila Babar and Natasha Noy and Nathan Byrd and Nick Roy and Nikola Momchev and Nilay Chauhan and Noveen Sachdeva and Oskar Bunyan and Pankil Botarda and Paul Caron and Paul Kishan Rubenstein and Phil Culliton and Philipp Schmid and Pier Giuseppe Sessa and Pingmei Xu and Piotr Stanczyk and Pouya Tafti and Rakesh Shivanna and Renjie Wu and Renke Pan and Reza Rokni and Rob Willoughby and Rohith Vallu and Ryan Mullins and Sammy Jerome and Sara Smoot and Sertan Girgin and Shariq Iqbal and Shashir Reddy and Shruti Sheth and Siim Põder and Sijal Bhatnagar and Sindhu Raghuram Panyam and Sivan Eiger and Susan Zhang and Tianqi Liu and Trevor Yacovone and Tyler Liechty and Uday Kalra and Utku Evci and Vedant Misra and Vincent Roseberry and Vlad Feinberg and Vlad Kolesnikov and Woohyun Han and Woosuk Kwon and Xi Chen and Yinlam Chow and Yuvein Zhu and Zichuan Wei and Zoltan Egyed and Victor Cotruta and Minh Giang and Phoebe Kirk and Anand Rao and Kat Black and Nabila Babar and Jessica Lo and Erica Moreira and Luiz Gustavo Martins and Omar Sanseviero and Lucas Gonzalez and Zach Gleicher and Tris Warkentin and Vahab Mirrokni and Evan Senter and Eli Collins and Joelle Barral and Zoubin Ghahramani and Raia Hadsell and Yossi Matias and D. Sculley and Slav Petrov and Noah Fiedel and Noam Shazeer and Oriol Vinyals and Jeff Dean and Demis Hassabis and Koray Kavukcuoglu and Clement Farabet and Elena Buchatskaya and Jean-Baptiste Alayrac and Rohan Anil and Dmitry and Lepikhin and Sebastian Borgeaud and Olivier Bachem and Armand Joulin and Alek Andreev and Cassidy Hardin and Robert Dadashi and Léonard Hussenot},
}

@misc{muennighoff2025s1simpletesttimescaling,
      title={s1: Simple test-time scaling}, 
      author={Niklas Muennighoff and Zitong Yang and Weijia Shi and Xiang Lisa Li and Li Fei-Fei and Hannaneh Hajishirzi and Luke Zettlemoyer and Percy Liang and Emmanuel Candès and Tatsunori Hashimoto},
      year={2025},
      eprint={2501.19393},
      archivePrefix={arXiv},
      primaryClass={cs.CL},
      url={https://arxiv.org/abs/2501.19393}, 
}

@inproceedings{
zheng2024lmsyschatm,
title={{LMSYS}-Chat-1M: A Large-Scale Real-World {LLM} Conversation Dataset},
author={Lianmin Zheng and Wei-Lin Chiang and Ying Sheng and Tianle Li and Siyuan Zhuang and Zhanghao Wu and Yonghao Zhuang and Zhuohan Li and Zi Lin and Eric Xing and Joseph E. Gonzalez and Ion Stoica and Hao Zhang},
booktitle={The Twelfth International Conference on Learning Representations},
year={2024},
url={https://openreview.net/forum?id=BOfDKxfwt0}
}

@article{ji2024emma500enhancingmassivelymultilingual,
      title={{EMMA}-500: Enhancing Massively Multilingual Adaptation of Large Language Models}, 
      author={Shaoxiong Ji and Zihao Li and Indraneil Paul and Jaakko Paavola and Peiqin Lin and Pinzhen Chen and Dayyán O'Brien and Hengyu Luo and Hinrich Schütze and Jörg Tiedemann and Barry Haddow},
      year={2024},
      journal={arXiv preprint 2409.17892},
      url={https://arxiv.org/abs/2409.17892}, 
}

@inproceedings{10.1145/3209978.3210080,
author = {Zhu, Yaoming and Lu, Sidi and Zheng, Lei and Guo, Jiaxian and Zhang, Weinan and Wang, Jun and Yu, Yong},
title = {Texygen: A Benchmarking Platform for Text Generation Models},
year = {2018},
isbn = {9781450356572},
publisher = {Association for Computing Machinery},
address = {New York, NY, USA},
url = {https://doi.org/10.1145/3209978.3210080},
doi = {10.1145/3209978.3210080},
booktitle = {The 41st International ACM SIGIR Conference on Research \& Development in Information Retrieval},
pages = {1097–1100},
numpages = {4},
location = {Ann Arbor, MI, USA},
series = {SIGIR '18}
}

@inproceedings{
shi2023language,
title={Language models are multilingual chain-of-thought reasoners},
author={Freda Shi and Mirac Suzgun and Markus Freitag and Xuezhi Wang and Suraj Srivats and Soroush Vosoughi and Hyung Won Chung and Yi Tay and Sebastian Ruder and Denny Zhou and Dipanjan Das and Jason Wei},
booktitle={The Eleventh International Conference on Learning Representations },
year={2023},
url={https://openreview.net/forum?id=fR3wGCk-IXp}
}

@inproceedings{
xu2024wizardlm,
title={Wizard{LM}: Empowering Large Pre-Trained Language Models to Follow Complex Instructions},
author={Can Xu and Qingfeng Sun and Kai Zheng and Xiubo Geng and Pu Zhao and Jiazhan Feng and Chongyang Tao and Qingwei Lin and Daxin Jiang},
booktitle={The Twelfth International Conference on Learning Representations},
year={2024},
url={https://openreview.net/forum?id=CfXh93NDgH}
}

@inproceedings{
zhou2023lima,
title={{LIMA}: Less Is More for Alignment},
author={Chunting Zhou and Pengfei Liu and Puxin Xu and Srini Iyer and Jiao Sun and Yuning Mao and Xuezhe Ma and Avia Efrat and Ping Yu and LILI YU and Susan Zhang and Gargi Ghosh and Mike Lewis and Luke Zettlemoyer and Omer Levy},
booktitle={Thirty-seventh Conference on Neural Information Processing Systems},
year={2023},
url={https://openreview.net/forum?id=KBMOKmX2he}
}

@misc{padmakumar2024doeswritinglanguagemodels,
      title={Does Writing with Language Models Reduce Content Diversity?}, 
      author={Vishakh Padmakumar and He He},
      year={2024},
      eprint={2309.05196},
      archivePrefix={arXiv},
      primaryClass={cs.CL},
      url={https://arxiv.org/abs/2309.05196}, 
}

@inproceedings{don-yehiya-etal-2025-sharelm,
    title = "The {S}hare{LM} Collection and Plugin: Contributing Human-Model Chats for the Benefit of the Community",
    author = "Don-Yehiya, Shachar  and
      Choshen, Leshem  and
      Abend, Omri",
    editor = "Mishra, Pushkar  and
      Muresan, Smaranda  and
      Yu, Tao",
    booktitle = "Proceedings of the 63rd Annual Meeting of the Association for Computational Linguistics (Volume 3: System Demonstrations)",
    month = jul,
    year = "2025",
    address = "Vienna, Austria",
    publisher = "Association for Computational Linguistics",
    url = "https://aclanthology.org/2025.acl-demo.17/",
    pages = "167--177",
    ISBN = "979-8-89176-253-4"
}

@misc{shaib2025standardizingmeasurementtextdiversity,
      title={Standardizing the Measurement of Text Diversity: A Tool and a Comparative Analysis of Scores}, 
      author={Chantal Shaib and Joe Barrow and Jiuding Sun and Alexa F. Siu and Byron C. Wallace and Ani Nenkova},
      year={2025},
      eprint={2403.00553},
      archivePrefix={arXiv},
      primaryClass={cs.CL},
      url={https://arxiv.org/abs/2403.00553}, 
}

@article{joulin2016bag,
  title={Bag of Tricks for Efficient Text Classification},
  author={Joulin, Armand and Grave, Edouard and Bojanowski, Piotr and Mikolov, Tomas},
  journal={arXiv preprint arXiv:1607.01759},
  year={2016}
}

@article{joulin2016fasttext,
  title={FastText.zip: Compressing text classification models},
  author={Joulin, Armand and Grave, Edouard and Bojanowski, Piotr and Douze, Matthijs and J{\'e}gou, H{\'e}rve and Mikolov, Tomas},
  journal={arXiv preprint arXiv:1612.03651},
  year={2016}
}
